\documentclass[10pt,journal,compsoc]{IEEEtran}
\ifCLASSOPTIONcompsoc
  \usepackage[nocompress]{cite}
\else
  \usepackage{cite}
\fi
\ifCLASSINFOpdf
\else
\fi

\usepackage{graphicx}
\usepackage{multirow}
\usepackage{booktabs}
\usepackage{amsmath,amssymb} 

\begin{document}
\title{Learning Structural Representations for Recipe Generation and Food Retrieval}

\author{Hao~Wang,
        Guosheng~Lin,
        Steven~C.~H.~Hoi,~\IEEEmembership{Fellow,~IEEE}
        and~Chunyan~Miao% <-this % stops a space
\IEEEcompsocitemizethanks{\IEEEcompsocthanksitem Hao Wang, Guosheng Lin and Chunyan Miao are with School of Computer Science and Engineering, Nanyang Technological University.
\hfil\break
E-mail: \{hao005,gslin,ascymiao\}@ntu.edu.sg.
\IEEEcompsocthanksitem Steven C. H. Hoi is with Singapore Management University. 
\hfil\break
E-mail: chhoi@smu.edu.sg.
}% <-this % stops an unwanted space
\IEEEcompsocitemizethanks{\IEEEcompsocthanksitem Corresponding authors: Chunyan Miao and Guosheng Lin.}
}


\markboth{Journal of \LaTeX\ Class Files,~Vol.~14, No.~8, August~2015}%
{Shell \MakeLowercase{\textit{et al.}}: Bare Demo of IEEEtran.cls for Computer Society Journals}



\IEEEtitleabstractindextext{%
\begin{abstract}
Food is significant to human daily life. In this paper, we are interested in learning structural representations for lengthy recipes, that can benefit the recipe generation and food cross-modal retrieval tasks. Different from the common vision-language data, here the food images contain mixed ingredients and target recipes are lengthy paragraphs, where we do not have annotations on structure information. To address the above limitations, we propose a novel method to unsupervisedly learn the sentence-level tree structures for the cooking recipes. Our approach brings together several novel ideas in a systematic framework: (1) exploiting an unsupervised learning approach to obtain the sentence-level tree structure labels before training; (2) generating trees of target recipes from images with the supervision of tree structure labels learned from (1); and (3) integrating the learned tree structures into the recipe generation and food cross-modal retrieval procedure. Our proposed model can produce good-quality sentence-level tree structures and coherent recipes. We achieve the state-of-the-art recipe generation and food cross-modal retrieval performance on the benchmark Recipe1M dataset.
\end{abstract}

}

\maketitle

\IEEEdisplaynontitleabstractindextext

% \fi
\IEEEpeerreviewmaketitle

\IEEEraisesectionheading{\section{Introduction}\label{sec:introduction}}
\IEEEPARstart{F}{ood}-related research with the newly evolved deep learning-based techniques is becoming a popular topic, as food is essential to human life. In this paper, we investigate the tasks of recipe generation \cite{salvador2019inverse} and food cross-modal retrieval \cite{salvador2017learning}, where we enhance their performance with the unsupervisedly learned recipe tree structures. To be specific, in the recipe generation task, we aim to generate the corresponding and coherent cooking instructions for the given food images with a language decoder. In the food cross-modal retrieval task, the goal is to retrieve the matched food images given recipes as the query, and vice versa. 

Both recipe generation and food cross-modal retrieval models aim to produce the corresponding recipes from given images. 
Since direct transforms from image to text would be difficult \cite{wang2016comprehensive}, many recent methods \cite{chen2019counterfactual,vinyals2015show,chun2021probabilistic} for image captioning and cross-modal retrieval adopt the detected object features to enhance image features. Moreover, learning scene graphs from the detected objects can boost the cross-modal task performance further \cite{gu2019unpaired,gu2018look}. This depicts the significance of using the graph information for the recipe generation and food cross-modal retrieval models.

In the food dataset Recipe1M \cite{salvador2017learning}, there have food images and the paired text annotations including ingredients and cooking instructions, where the food images are static and contain all the mixed ingredients. Generally, the task settings of the recipe generation and food cross-modal retrieval are almost the same as those of image captioning and general cross-modal retrieval. However, there still exist two differences between them: (i) the given text length and (ii) the annotations on the data structural information.

First, most popular image-text cross-modal datasets, such as Flickr \cite{plummer2015flickr30k} and MS-COCO \cite{chen2015microsoft} datasets, only have one sentence per caption. By contrast, cooking instructions are paragraphs, containing multiple sentences to represent the cooking process, which cannot be fully shown in a single food image. 
Therefore, generating lengthy recipes with traditional image captioning models \cite{vinyals2015show,chen2019counterfactual,gu2019unpaired} may hardly capture the whole cooking procedure. Second, the lack of structural information labeling is another challenge in our researched food cross-modal tasks. For example, MS-COCO \cite{chen2015microsoft} has precise bounding box annotations in images, giving scene graph information for the caption generation or the retrieval process. This structural information makes it easier to represent the inner relationships of the image or text data. While in food images, all the ingredients are mixed when cooked, making it difficult to obtain the detection labeling in the food images. Moreover, there are no existing works attempting to learn the structure representations for the cooking recipes. To address this limitation, we propose to learn the recipe tree structures in an unsupervised manner, integrate the learned trees into the recipe generation and food cross-modal retrieval tasks and improve their performance.

\begin{figure*}[t]
\begin{center}
\includegraphics[width=0.8\textwidth]{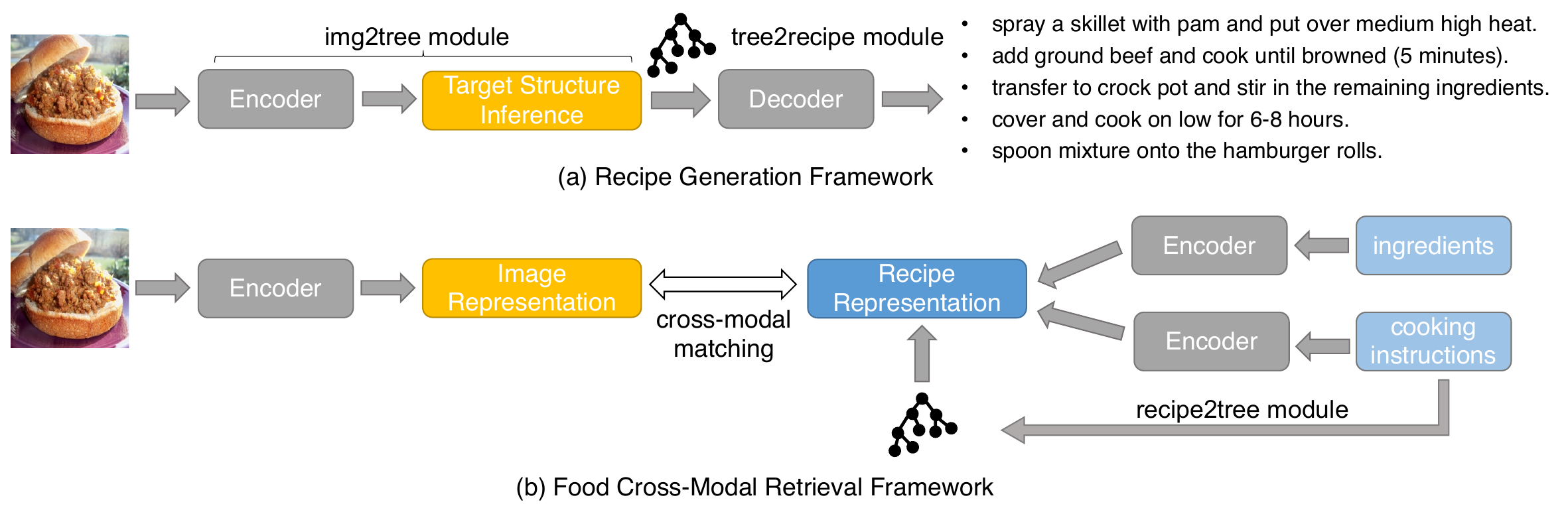}
\end{center}
\vspace{-0.25in}
   \caption{The demonstration of our proposed frameworks. In the recipe generation model (a), we use the img2tree module to infer the recipe tree structures, which are encoded with the graph attention networks to give tree embeddings in the tree2recipe module. Then we generate the recipes. In the food cross-modal retrieval framework (b), we incorporate the tree structures from the recipe2tree module into the original recipe features, which boosts food cross-modal matching performance.
}
\label{fig:demo}
% \vspace{-0.3in}
\end{figure*}

Specifically, benefiting from the recent advances in the language parsing, some research works, such as ON-LSTM \cite{shen2018ordered}, utilize an unsupervised way to produce word-level parsing trees of sentences and achieve good results. Inspired by that, we propose a novel recipe2tree module, where we extend the ON-LSTM architecture to do sentence-level tree structure generation. We propose to train the extended ON-LSTM with the Quick Thoughts manner \cite{logeswaran2018efficient}, to capture the order information inside recipes. By doing so, we get the recipe tree structure labels.

We then apply the obtained recipe structure information on the recipe generation and food cross-modal retrieval tasks. For the recipe generation, we propose a novel framework named \textbf{S}tructure-aware \textbf{G}eneration \textbf{N}etwork (SGN) to integrate the tree structure information into the training and inference phases, as shown in Figure \ref{fig:demo}(a). SGN is implemented to add a target structure inference module on the recipe generation process. Specifically, we propose to use a RNN to generate the recipe tree structures from food images. Based on the generated trees, we adopt the graph attention networks (GAT) to embed the trees, in an attempt to giving the model more guidance when generating recipes. With the tree structure embeddings, we make the generated recipes remain long-length as the ground truth, and improve the generation performance considerably.

We further demonstrate the efficacy of our unsupervisedly-learned recipe tree structures in the food cross-modal retrieval task, where we incorporate the tree representations into the recipe features, as shown in Figure \ref{fig:demo}(b). Specifically, we first produce the tree structures from the cooking instructions with our proposed recipe2tree module. Then we also use the GAT to give the tree features, which are adopted to fuse with the recipe representations. We conduct the cross-modal matching between the enhanced image and recipe representations.

Our contributions can be summarized as:
% \vspace{-0.05in}
\begin{itemize}
   \item We propose a recipe2tree module to capture latent sentence-level tree structures for recipes, which are learned through an unsupervised approach. The obtained tree structures are adopted to supervise the following img2tree module. 
   \item We propose to use the img2tree module to generate the recipe tree structures from food images, where we use a RNN for conditional tree generation.
   \item We propose to utilize the tree2recipe module, which encodes the inferred tree structures. It is implemented with graph attention networks, and boosts the recipe generation performance.
   \item We show the tree structures learned in the recipe2tree module can also help on improving the cross-modal retrieval performance.  
\end{itemize}
% \vspace{-0.05in}

Figure \ref{fig:demo} shows the concise demonstration on our proposed frameworks for the recipe generation and food cross-modal retrieval tasks. We have conducted extensive experiments to evaluate the recipe generation and food retrieval performance, showing our proposed method outperforms state-of-the-art baselines on Recipe1M dataset \cite{salvador2017learning}. We also present ablation studies as well as the qualitative results of the recipe generation and food cross-modal retrieval results.

Our preliminary research has been published in \cite{wang2020structure}. The code is publicly available\footnote{https://github.com/hwang1996/SGN}.

\section{Related Work}

\subsection{Image captioning}
Image captioning task is defined as generating the corresponding text descriptions from images. Based on MS-COCO dataset \cite{chen2015microsoft}, most existing image captioning techniques adopt deep learning-based model. One popular approach is Encoder-Decoder architecture \cite{vinyals2015show,bahdanau2014neural,golland2010game,kazemzadeh2014referitgame}, where a CNN is used to obtain the image features along with object detection, then a language model is used to convert the image features into text. 

Since image features are fed only at the beginning stage of generation process, the language model may face vanishing gradient problem \cite{hossain2019comprehensive}. Therefore, image captioning model is facing challenges in long sentence generation \cite{bahdanau2014neural}. To enhance text generation process, \cite{chen2019counterfactual,gu2019unpaired} involve scene graph into the framework. However, scene graph generation rely heavily on object bounding box labeling, which is provided by MS-COCO dataset. When we shift to some other datasets without rich annotation, we can hardly obtain the graph structure information of the target text. Meanwhile, crowdsourcing annotation is high-cost and may not be reliable. Therefore, we propose to produce tree structures for paragraphs unsupervisedly, helping the recipe generation task in Recipe1M dataset \cite{salvador2017learning}. 

\subsection{Multimodal food computing}
Food computing \cite{min2019survey} has raised great interest recently, it targets applying computational approaches for analyzing multimodal food data for recognition \cite{bossard2014food}, retrieval \cite{salvador2017learning,wang2019learning,wang2021cross} and generation \cite{salvador2019inverse} of food. In this paper, we choose Recipe1M dataset \cite{salvador2017learning} to validate our proposed method on recipe generation and food cross-modal retrieval task. 

Recipe generation is a challenging task, it is mainly because that recipes (cooking instructions) contain multiple sentences. Salvador et al. \cite{salvador2019inverse} adopt transformer to generate lengthy recipes, but they fail to consider the holistic recipe structure before generation, hence their generated recipes may miss some steps. In contrast, our proposed method allows the model to predict the recipe tree structures first, and then give better generation results. Food cross-modal retrieval targets retrieving matched items given one food image or recipe. Prior works \cite{wang2019learning,chen2018deep,salvador2017learning,carvalho2018cross} mainly aim to align the cross-modal embeddings in the common space, we improve the retrieval baseline results by enhancing the recipe representations with learned tree structures.

\begin{figure*}[t]
\begin{center}
\includegraphics[width=0.7\textwidth]{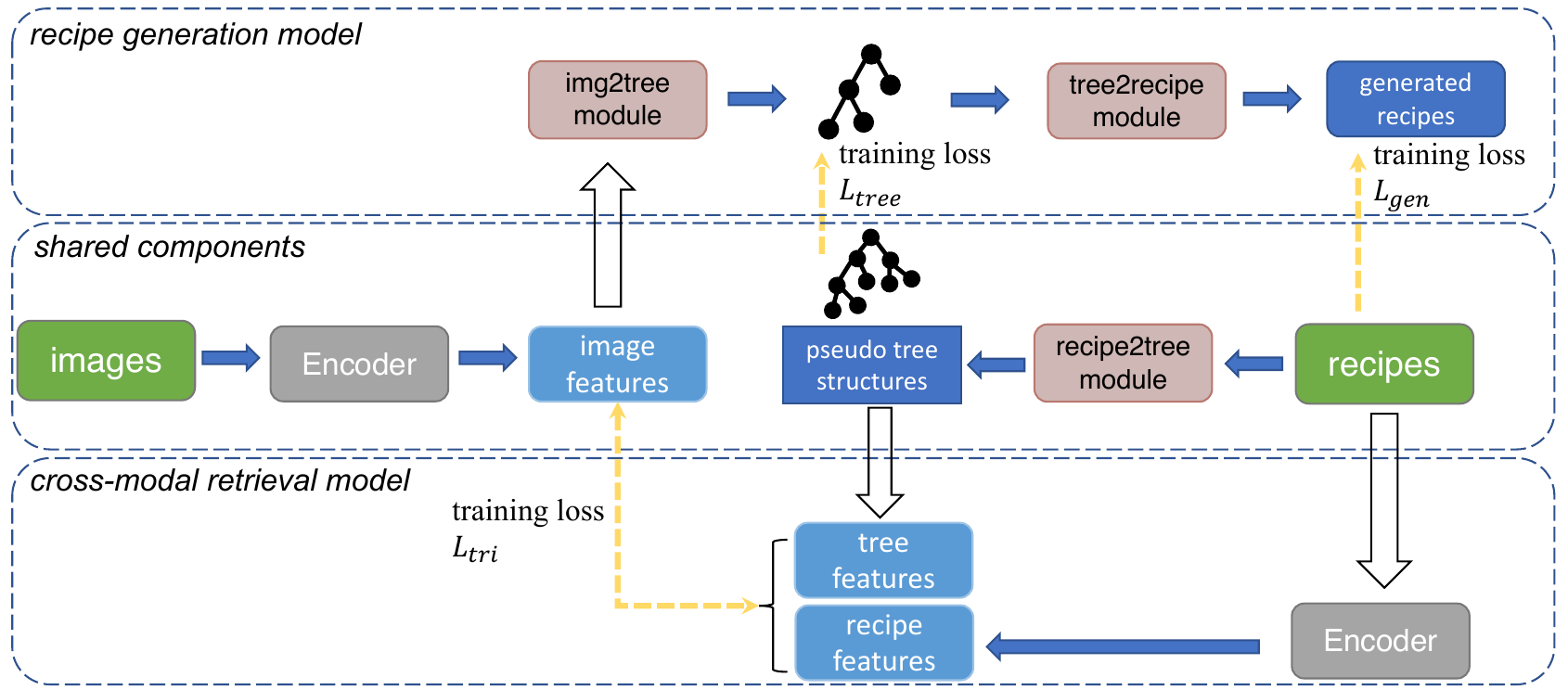}
\end{center}
\vspace{-0.2in}
   \caption{The concise training flow of our proposed models for recipe generation and food cross-modal retrieval, which are trained individually. In the shared components, we extract the food image features and use the recipe2tree module to produce pseudo recipe tree structures. In the recipe generation model, we generate trees with the img2tree module that is trained with $L_{tree}$. The predicted tree structures from the img2tree module are used to generate the recipes in the tree2recipe module, which is supervised by $L_{gen}$. In the cross-modal retrieval model, we concatenate the tree features and recipe features. The concatenated features are trained to match with the image features, which is supervised by $L_{tri}$.}
\label{fig:training}
% \vspace{-0.2in}
\end{figure*}

\subsection{Image-to-text retrieval}
The image-to-text retrieval task is to retrieve the corresponding image given the text, and vice versa. Prevailing methods \cite{salvador2017learning,wang2021cross,vo2019composing,chen2020image} adopt the deep neural networks to give the image and text features respectively, and use the metric learning to map the cross-modal features into a common space, such that the alignment between the text and images can be achieved. Specifically, Vo et al. \cite{vo2019composing} utilize the image plus some text to retrieve the images with certain language attributes. They propose to combine image and text through residual connection and produce the image-text joint features to do the retrieval task. Chen et al. \cite{chen2020image} conduct experiments with the same setting as \cite{vo2019composing}, where they use a composite transformer to plug in a CNN and then selectively preserve and transform the visual features conditioned on language semantics. In the domain of food cross-modal retrieval, Salvador et al. \cite{salvador2017learning} aim to learn joint embeddings (JE) for images and recipes, where they adopt cosine loss to align image-recipe pairs and classification loss to regularize the learning. Zhu et al. \cite{zhu2019r2gan} use two-level ranking loss at embedding and image spaces in $\mathbf{R^2 GAN}$. Wang et al. \cite{wang2019learning} introduce the translation consistency component to allow feature distributions from different modalities to be similar.

\subsection{Language parsing}
Parsing is served as one effective language analysis tool, it can output the tree structure of a string of symbols. Generally, language parsing is divided into word-level and sentence-level parsing. Word-level parsing is also known as grammar induction, which aims at learning the syntactic tree structure from corpora data. 
Some of the research works use a supervised way to predict the corresponding latent tree structure given a sentence \cite{wu2017sequence,socher2013recursive}. However, precise parser annotation is hard to obtain. \cite{williams2017broad,shen2017neural,shen2018ordered} explored to learn the latent structure without the expert-labeled data. 
Especially, Shen et al. \cite{shen2018ordered} propose to use ON-LSTM, which equips the LSTM architecture with an inductive bias towards learning latent tree structures. They train the model with normal language modeling way, at the same time they can get the parsing output induced by the model.

Sentence-level parsing is used to identify the elementary discourse units in a text, and it brings some benefits to discourse analysis. Many recent works attempted to use complex model with labeled data to achieve the goal \cite{jia2018modeling,lin2019unified}. Here we extend ON-LSTM \cite{shen2018ordered} for unsupervised sentence-level parsing, which is trained using quick thoughts \cite{logeswaran2018efficient}.

\subsection{Graph generation}
Graph is the natural and fundamental data structure in many fields, such as social networks and biology, and a tree is an undirected graph. 
The basic idea of graph generation model is to make auto-regressive decisions during graph generation. For example, Li et al. \cite{li2018learning} add graph nodes and edges sequentially with auto-regressive models. The tree generation approach we use is similar with GraphRNN \cite{you2018graphrnn}. They \cite{you2018graphrnn} first map the graph to sequence under random ordering, then use edge-level and graph-level RNN to update the adjacency vector. While in our tree generation method, we generate the tree conditioned on food images, and the node ordering is fixed according to hierarchy, which releases the complexity of the sampling space.

\section{Method}
Here we investigate two research tasks of 1) food recipe generation from images and 2) food cross-modal retrieval. We give the unified training flow demonstration for these two tasks in Figure \ref{fig:training}. In the following sections, we present our proposed models for recipe generation and food cross-modal retrieval, whose frameworks are shown in Figure \ref{fig:pipeline} and Figure \ref{fig:retrieval} respectively.

\subsection{Overview}

We show the concise training flows for recipe generation and food cross-modal retrieval models in Figure \ref{fig:training}, where we first use image encoders to extract visual features and use the recipe2tree module to give recipe tree structures. Specifically, adopting the generated pseudo structures of the recipe2tree module is the core component of recipe generation and cross-modal retrieval models, which is shared by both models. We learn recipe sentence-level tree structures in an unsupervised manner at the recipe2tree module. Then in the recipe generation model, the generated pseudo trees are used to supervise the tree generation from images; in the cross-modal retrieval model, the tree structures are adopted to enhance the recipe language features.

Technically, in the recipe2tree module we propose to use the hierarchical ON-LSTM \cite{shen2018ordered} to encode the cooking instructions and train the ON-LSTM with quick thoughts approach \cite{logeswaran2018efficient}. Then we can obtain the latent tree structures of cooking instructions, which are used as the pseudo labels to supervise the training of the img2tree module.

\begin{figure*}
\begin{center}
\includegraphics[width=0.75\textwidth]{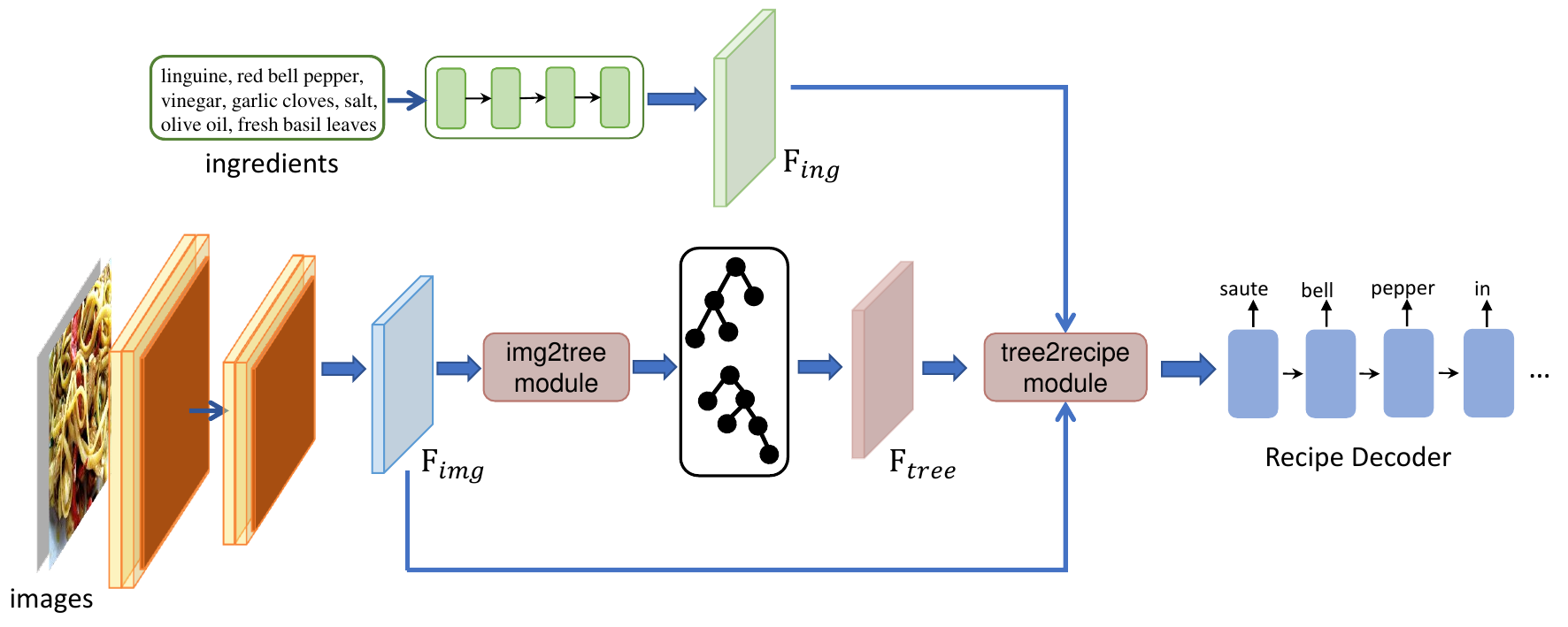}
\end{center}
\vspace{-0.2in}
   \caption{Our proposed framework for recipe generation. The ingredients and food images are embedded by a pretrained language model and CNN respectively to produce the output features $\mathrm{F}_{ing}$ and $\mathrm{F}_{img}$. Before language generation, we first infer the tree structure of target cooking instructions. To do so, we utilize the img2tree module, where a RNN produces the nodes and edge links step-by-step based on $\mathrm{F}_{img}$. Then in tree2recipe module, we adopt graph attention networks (GAT) to encode the generated tree adjacency matrix, and get tree embedding $\mathrm{F}_{tree}$. We combine $\mathrm{F}_{ing}$, $\mathrm{F}_{img}$ and $\mathrm{F}_{tree}$ to construct a final embedding for recipe generation, which is performed using a transformer.}
\label{fig:pipeline}
% \vspace{-0.1in}
\end{figure*}

During the recipe generation training phase, as shown in Figures \ref{fig:training} and \ref{fig:pipeline}, we input food images and ingredients to our proposed model. We try two different language models to encode ingredients, i.e. non-pretrained and pretrained model, to get the ingredient features $\mathrm{F}_{ing}$. In the non-pretrained model training, we use one word embedding layer \cite{salvador2019inverse} to give $\mathrm{F}_{ing}$. Besides, we adopt BERT \cite{devlin2018bert} for ingredient embedding, which is one of the state-of-the-art NLP pretrained models. In the image embedding branch, we adopt a CNN to encode the food images and get the image features $\mathrm{F}_{img}$. Based on $\mathrm{F}_{img}$, we generate the sentence-level tree structures and make them align with the pseudo labels produced by the recipe2tree module. Specifically, we transform the tree structures to a $1$-dimensional adjacency sequence for RNN to generate, where the RNN's initial state is image feature $\mathrm{F}_{img}$. To incorporate the generated tree structure into the recipe generation process, we get the tree embedding $\mathrm{F}_{tree}$ with graph attention networks (GAT) \cite{velivckovic2017graph}, and concatenate it with the image features $\mathrm{F}_{img}$ and ingredient features $\mathrm{F}_{ing}$. We then generate the recipes conditioned on the concatenated features of $\langle \mathrm{F}_{tree}, \mathrm{F}_{img}, \mathrm{F}_{ing} \rangle$ with a transformer \cite{vaswani2017attention}.

Our proposed recipe generation framework is optimized over two objectives: to generate reasonable recipes given the food images and ingredients; and to produce the sentence-level tree structures of target recipes. The overall
objective is given as:
% \vspace{-0.1in}
\begin{equation}
% \vspace{-0.1in}
\label{eq:obj}
    L = \lambda_1L_{gen} + \lambda_2L_{tree},
\end{equation}
where $\lambda_1$ and $\lambda_2$ are trade-off parameters. $L_{gen}$ controls the recipe generation training with the input of $\langle \mathrm{F}_{tree}, \mathrm{F}_{img}, \mathrm{F}_{ing} \rangle$, and outputs the probabilities of word tokens. $L_{tree}$ is the tree generation loss, supervising the img2tree module to generate trees from images.

As shown in Figures \ref{fig:training} and \ref{fig:retrieval}, we propose to incorporate the latent trees into the food cross-modal retrieval task to further demonstrate the usefulness of our unsupervisedly-learned tree structures and boost the retrieval performance. In this task, given a food image, we want to retrieve the corresponding cooking recipe including the ingredients and the cooking instructions, or vice versa. 
To this end, we also adopt a CNN to give image representations $\mathrm{F}_{img}$ and a language encoder to give the ingredient and cooking instruction features $\mathrm{F}_{ing}$ and $\mathrm{F}_{ins}$. We adopt the GAT to encode the sentence-level tree structures and obtain the tree representations $\mathrm{F}_{tree}$. The recipe embeddings $\mathrm{F}_{rec}$ are constructed by the concatenation of $\langle \mathrm{F}_{tree}, \mathrm{F}_{ins}, \mathrm{F}_{ing} \rangle$. We use triplet loss $L_{tri}$ to align image and recipe representations $\mathrm{F}_{img}$ and $\mathrm{F}_{rec}$ in the feature space and learn a joint embedding for cross-modal matching.

\subsection{ON-LSTM revisit}
Ordered Neurons LSTM (ON-LSTM) \cite{shen2018ordered} is proposed to infer the underlying tree-like structure of language while learning the word representation. It can achieve good performance in unsupervised parsing task. ON-LSTM is constructed based on the intuition that each node in the tree can be represented by a set of neurons   the hidden states of recurrent neural networks. To this end, ordered neuron is an inductive bias, where high-ranking neurons store long-term information, while low-ranking neurons contain short-term information that can be rapidly forgotten. Instead of acting independently on each neuron, the gates of ON-LSTM are dependent on the others by enforcing the order in which neurons should be updated. Technically, Shen et al. \cite{shen2018ordered} define the split point $d$ between two segments. $d^f_t$ and $d^i_t$ represent the hierarchy of the previous hidden states $h_{t-1}$ and that of the current input token $x_t$ respectively, which can be formulated as:
\begin{align}
    d^f_t &= \mathrm{softmax} (W_{\Tilde{f}}x_{t}+U_{\Tilde{f}}h_{t-1}+b_{\Tilde{f}}), \label{eq:masterforget} \\
    d^i_t &= \mathrm{softmax} (W_{\Tilde{i}}x_{t}+U_{\Tilde{i}}h_{t-1}+b_{\Tilde{i}}), \label{eq:masterinput}
\end{align}
where $\Tilde{f}$ and $\Tilde{i}$ are defined by the ON-LSTM as the master forget gate and the master input gate. $W$, $U$ and $b$ are the learnable weights of ON-LSTM. As stated in \cite{shen2018ordered}, the information stored in the first $d^f_t$ neurons of the previous cell state will be completely erased, and a large $d^i_t$ means that the current input $x_t$ contains long-term information that needs to be preserved for several time steps. The model weights are updated based on the predicted $\Tilde{f}$ and $\Tilde{i}$. 

The ON-LSTM model is trained through word-level language modeling, where given one token in the document they predict the next token. With the trained ON-LSTM, Shen et al. attempt to do the unsupervised constituency parsing. At each time step, they compute an estimate of $d^f_t$:
\begin{equation}
    \hat{d}^f_t = \mathbb{E} \left[ d^f_t \right] = \sum_{k=1}^{D_m} k p_f(d_t=k),
\end{equation}
where $p_f$ denotes the probability distribution over split points associated to the master forget gate and $D_m$ is the size of the hidden state. Given $\hat{d}^f_t$, the top-down greedy parsing algorithm \cite{shen2017neural} is used for unsupervised constituency parsing. As described in \cite{shen2018ordered}, for the first $\hat{d}^f_i$, they split the sentence into constituents $((x_{<i}), (x_{i}, (x_{>i})))$. Then, they recursively repeat this operation for constituents $(x_{<i})$ and $(x_{>i})$, until each constituent contains only one word.
 
Therefore, ON-LSTM is able to discern a hierarchy between words based on the model neurons. However, ON-LSTM is originally trained by language modeling way and learns the word-level order information. To unsupervisedly produce sentence-level tree structure, we extend ON-LSTM in the recipe2tree module.

\subsection{Recipe2tree module} \label{sec:r2t}
In this module, we propose to learn a hierarchical ON-LSTM, i.e. word-level and sentence-level ON-LSTM. Specifically, in the word-level ON-LSTM, we input the cooking recipe word tokens and use the output features as the sentence embeddings. The sentence embeddings will be fed into the sentence-level ON-LSTM for end-to-end training. 

Since the original training way \cite{shen2018ordered}, such as language modeling or seq2seq \cite{britz2017massive} word prediction training, cannot be used in sentence representation learning, we incorporate the idea of quick thoughts (QT) \cite{logeswaran2018efficient} to supervise the hierarchical ON-LSTM training. The general objective of QT is a discriminative approximation where the model attempts to identify the embedding of a correct target sentence given a set of sentence candidates. In other words, instead of predicting \emph{what is the next} in language modeling, we predict \emph{which is the next} in QT training to capture the order information inside recipes. Technically, for each recipe data, we select first $N-1$ of the cooking instruction sentences as context, i.e. $S_{ctxt} = \{s_{1}, ..., s_{N-1}\}$. Then sentence $s_N$ turns out to be the correct next one. Besides, we randomly select $K$ sentences along with the correct sentence $s_N$ from each recipe, to construct candidate sentence set $S_{cand} = \{s_N, s_i, ..., s_k\}$. The candidate sentence features $g(S_{cand})$ are generated by the word-level ON-LSTM, and the context embeddings $f(S_{ctxt})$ are obtained from the sentence-level ON-LSTM. The computation of probability is given by
\begin{equation}
	p(s_{\text{cand}}|S_\text{ctxt}, S_\text{cand}) = \frac{\text{exp}[c(f(S_\text{ctxt}), g(s_\text{cand}))]}{\sum_{s'\in S_\text{cand}}\text{exp}[c(f(S_\text{ctxt}), g(s'))]},
\end{equation}
where $c$ is an inner product, to avoid the model learning poor sentence encoders and a rich classifier. Minimizing the number of parameters in the classifier encourages the encoders to learn disentangled and useful representations \cite{logeswaran2018efficient}. The training objective maximizes the probability of identifying the correct next sentences for each training recipe data $D$:
% % \vspace{-0.1in}
\begin{equation}
% \vspace{-0.1in}
	\sum_{s \in D} \text{log } p(s |S_\text{ctxt}, S_\text{cand}).
\end{equation}

We adopt the learned sentence-level ON-LSTM to give the neuron ranking for the cooking instruction sentences, which can be converted to the recipe tree structures $T$ through the top-down greedy parsing algorithm \cite{shen2017neural}. $T$ are adopted as the pseudo labels to supervise the training of img2tree module.

\subsection{Recipe generation}
\subsubsection{Img2tree module}
In the img2tree module, we aim to generate the tree structures from food images. Tree structure has hierarchical nature, in other words, ``parent" node is always one step higher in the hierarchy than ``child'' nodes. Given the properties, we first represent the trees as sequence under the hierarchical ordering. Then, we use an auto-regressive model to model the sequence, meaning that the edges between subsequent nodes are dependent on the previous ``parent" node. Besides, in Recipe1M dataset, the longest cooking instructions have $19$ sentences. Therefore, the sentence-level parsing trees have limited node numbers, which avoids the model generating too long or complex sequence.

In Figure \ref{fig:pipeline}, we specify our tree generation approach. The generation process is conditioned on the food images. According to the hierarchical ordering, we first map the tree structure to the adjacency matrix, which denotes the links between nodes by $0$ or $1$. Then the lower triangular part of the adjacency matrix will be converted to a vector $V \in \mathbb{R}^{n\times1}$, where each element $V_i \in {\{0, 1\}}^i, i \in \{1, \dots, n\}$. Since edges in tree structure are undirected, $V$ can determine a unique tree $T$. 

Here the tree generation model is built based on the food images, capturing how previous nodes are interconnected and how following nodes construct edges linking previous nodes. Hence, we adopt Recurrent Neural Networks (RNN) to model the predefined sequence $V$. We use the image encoded features $\mathrm{F}_{img}$ as the initialization of RNN hidden state, and the state-transition function $h$ and the output function $y$ are formulated as:
% \vspace{-0.1in}
\begin{equation}
% \vspace{-0.1in}
	h_0 = \mathrm{F}_{img}, h_i = f_{trans}(h_{i-1}, V_{i-1}),
\end{equation}
\begin{equation}
% \vspace{-0.1in}
	y_i = f_{out}(h_{i}),
\end{equation}
where $h_i$ is conditioned on the previous generated $i-1$ nodes, $y_i$ outputs the probabilities of next node's adjacency vector. 

The tree generation objective function is:
% \vspace{-0.1in}
\begin{equation}
% \vspace{-0.1in}
    p(V) = \prod_{i=1}^{n}p(V_i |V_1, \dots, V_{i-1}),
\end{equation}
\begin{equation}
% \vspace{-0.1in}
    L_{tree} = \sum_{V \in D} \text{log } p(V),
\end{equation}
where $p(V)$ is the product of conditional distributions over the elements, $D$ denotes all the training data. 

\begin{figure}
\begin{center}
\includegraphics[width=0.25\textwidth]{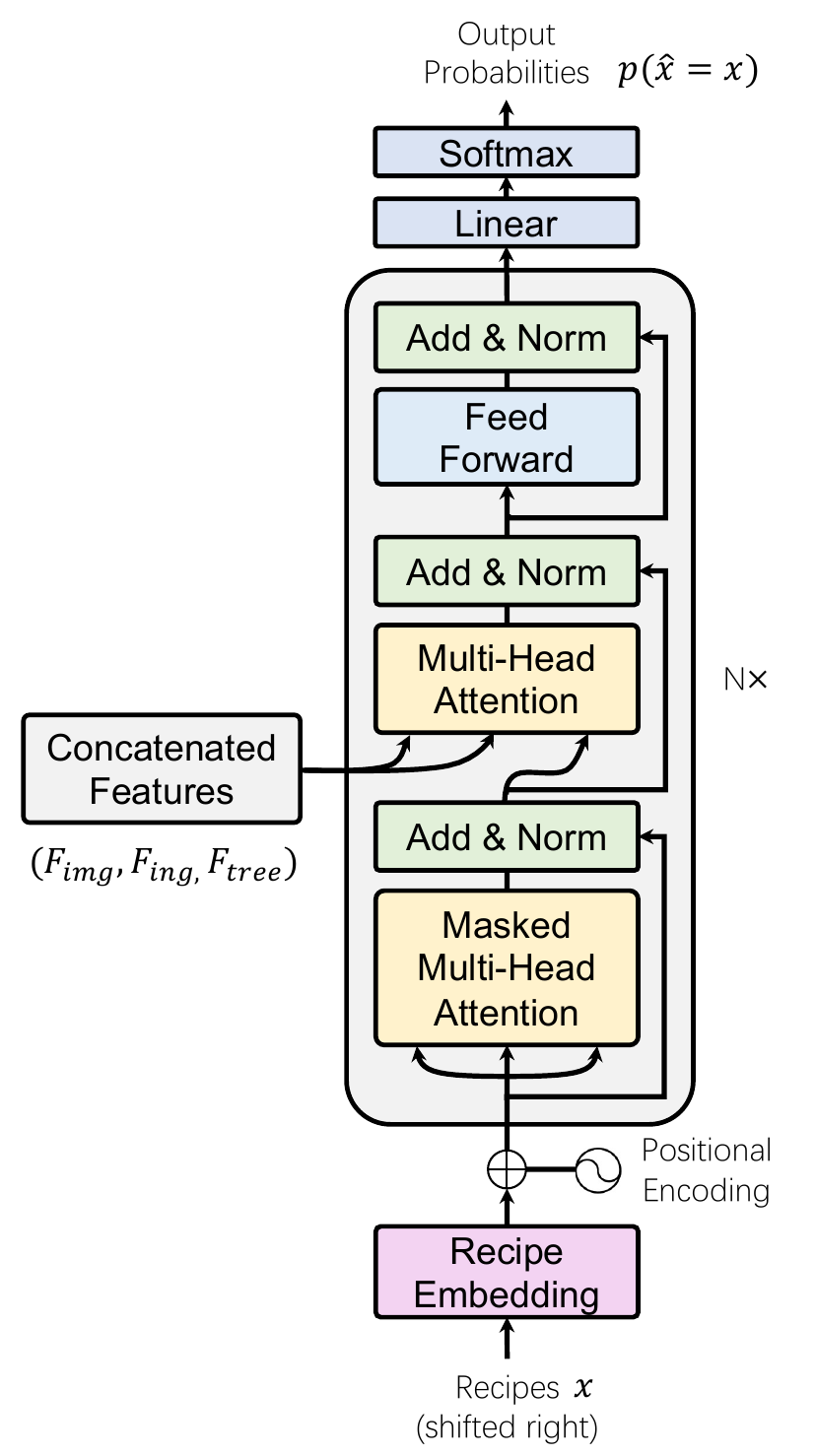}
\end{center}
% \vspace{-0.2in}
   \caption{The demonstration of the transformer training for the recipe generation. The concatenated features are composed of the image features $\mathrm{F}_{img}$, ingredient features $\mathrm{F}_{ing}$ and tree structure representations $\mathrm{F}_{tree}$. In the training phase, we take the ground truth recipes $x$ as the input. The predicted token and the output probability are denoted as $\hat{x}$ and $p(\hat{x}=x)$ respectively. We set the transformer layer number $N=16$.
}
\label{fig:trans}
% \vspace{-0.2in}
\end{figure}

\begin{figure*}[t]
\begin{center}
\includegraphics[width=0.65\textwidth]{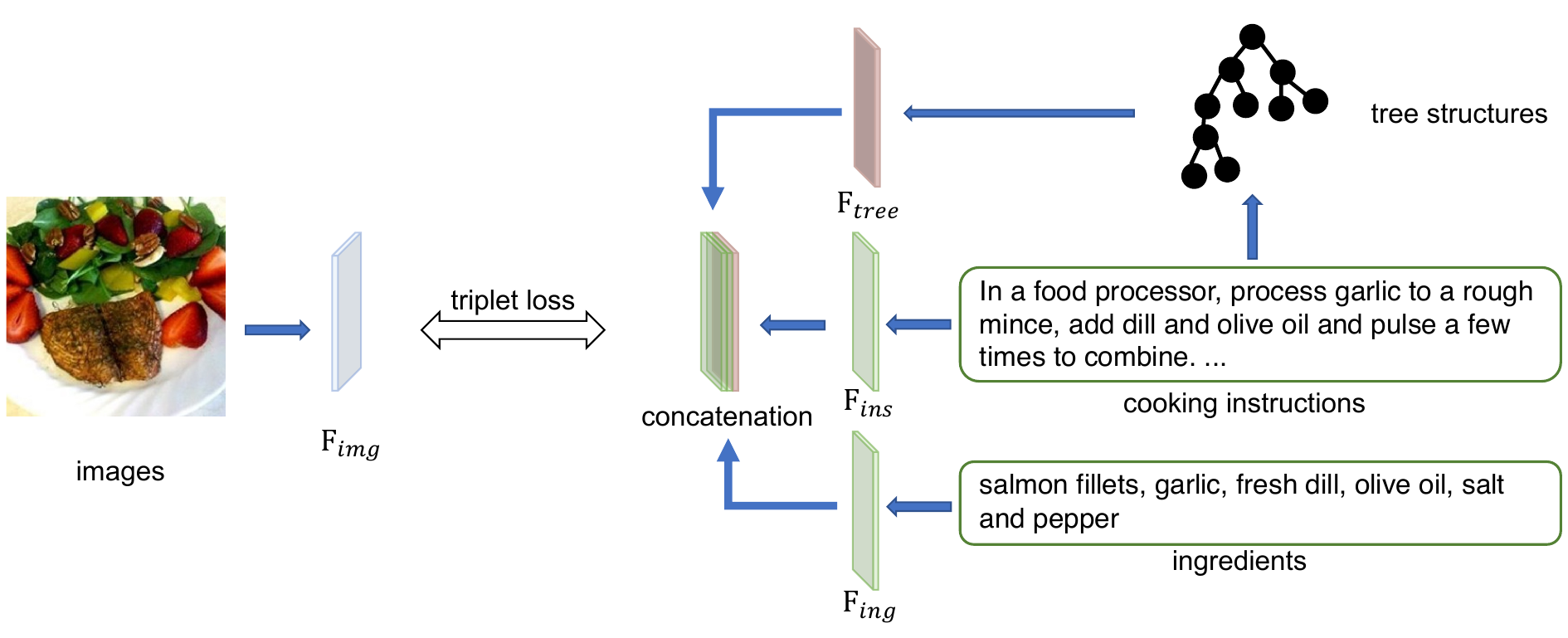}
\end{center}
\vspace{-0.3in}
   \caption{The training flow of food cross-modal retrieval. We first produce the sentence-level tree structures from the cooking instructions, where we use the sentence features as the node features. The tree structure, cooking instruction and ingredient features are denoted as $\mathrm{F}_{tree}$, $\mathrm{F}_{ins}$ and $\mathrm{F}_{ing}$ respectively. We use the concatenation of $\mathrm{F}_{tree}$, $\mathrm{F}_{ins}$ and $\mathrm{F}_{ing}$ as the recipe features. The triplet loss is adopted to learn the similarity between the recipe features $\mathrm{F}_{rec}$ and image features $\mathrm{F}_{img}$.}
\label{fig:retrieval}
% \vspace{-0.2in}
\end{figure*}

\subsubsection{Tree2recipe module}\label{sec:tree2re}
In the tree2recipe module, we utilize graph attention networks (GAT) \cite{velivckovic2017graph} to encode the generated trees. 
The input of GAT is the generated sentence-level tree adjacency matrix $A$ and its node features. Since the sentence features are not available during recipe generation, we produce node features with a linear transformation $\mathrm{W}$, which is applied on the adjacency matrix $A$. We then perform attention mechanism on the between connected nodes $(z_i, z_j)$ and compute the attention coefficients
% \vspace{-0.1in}
\begin{equation} \label{eq: attention}
% \vspace{-0.1in}
    % e_{ij} = \mathrm{attention}(\mathrm{W}z_i, \mathrm{W}z_j)
    e_{ij} = (\mathrm{W}z_i) {(\mathrm{W}z_j)}^T,
\end{equation}
where $e_{ij}$ measures the importance of node $j$'s features to node $i$, the attention coefficients are computed by the matrix multiplication. 

It is notable that different from most attention mechanism, where every node attends on every other node, GAT only allows each node to attend on its neighbour nodes. The underlying reason is that doing global attention fails to consider the property of tree structure, that each node has limited links to others. While the local attention mechanism used in GAT preserves the structural information well. We can formulate the final attentional score as:
% \vspace{-0.05in}
\begin{equation} \label{eq: alpha}
% \vspace{-0.1in}
    \alpha_{ij} = \mathrm{softmax}_j(e_{ij}) = \frac{\mathrm{exp}(e_{ij})}  
    {\sum_{k \in \mathrm{N}_i} \mathrm{exp}(e_{ik})},
\end{equation}
where $\mathrm{N}_i$ is the neighborhood of node $i$, the output score is normalized through the softmax function. Similar with \cite{vaswani2017attention}, GAT employs multi-head attention and averaging to stabilize the learning process. We get the tree features by the product of the attentional scores and the node features, and we perform nonlinear activation $\sigma$ on the output to get the final features:
% \vspace{-0.1in}
\begin{equation}
% \vspace{-0.1in}
    \mathrm{F}_{tree} = \sigma(\sum_{j \in \mathrm{N}_i} \alpha_{ij}\mathrm{W}z_j).
\end{equation}

\subsubsection{Recipe generation from images}
The demonstration of the transformer \cite{vaswani2017attention} structure for language generation is presented in Figure \ref{fig:trans}.
We adopt a 16-layer transformer \cite{vaswani2017attention} for recipe generation, which is the same setting as \cite{salvador2019inverse}. We use the teacher forcing training strategy, where we feed the previous ground truth word $x^{(i-1)}$ into the model and let the model generate the next word token $\hat{x}^{(i)}$ in the training phase.
In the transformer attention mechanism \cite{vaswani2017attention}, we have query $Q$, key $K$ and value $V$, the attentional output can be computed as
\begin{equation}
\mathrm{Attention}(Q, K, V) = \mathrm{softmax}(\frac{QK^T}{\sqrt{d_k}})V,
\end{equation}
where $d_k$ denotes the dimension of $K$. Here in the Multi-Head Attention module of Figure \ref{fig:trans}, we use the concatenated features of previously obtained $\mathrm{F}_{img}$, $\mathrm{F}_{ing}$ and $\mathrm{F}_{tree}$ as $K$ and $V$, and the processed recipe embeddings are used as the $Q$ . 

The training objective of the recipe generation is to maximize the following objective:
% \vspace{-0.1in}
\begin{equation}
% \vspace{-0.1in}
\label{objective}
    L_{gen} = \sum_{i=0}^{M} \text{log } p(\hat{x}^{(i)} = x^{(i)}),
\end{equation}
where $L_{gen}$ is the recipe generation loss, and $M$ is the maximum sentence generation length, $x^{(i)}$ and $\hat{x}^{(i)}$ denote the ground truth and generated tokens respectively. In the inference phase, the transformer decoder outputs $\hat{x}^{(i)}$ one by one.

\subsection{Food cross-modal retrieval}
The training framework for the food cross-modal retrieval task is shown in Figure \ref{fig:retrieval}.
We follow the same food cross-modal retrieval setting as \cite{salvador2017learning,wang2019learning,wang2021cross}, where given a food image we aim to find the corresponding cooking recipe, and vice versa. To this end, we first obtain the feature representations of food images and recipes respectively, then we learn the similarity between the food images and cooking recipes through the triplet loss. Technically, we get the food image representations $\mathrm{F}_{img}$ from the output of CNN directly, and get the recipe representations $\mathrm{F}_{rec}$ from the concatenation of the ingredient features $\mathrm{F}_{ing}$, instruction features $\mathrm{F}_{ins}$ and recipe tree structure representation $\mathrm{F}_{tree}$. We project $\mathrm{F}_{img}$ and $\mathrm{F}_{rec}$ into a common space and align them to realize cross-modal retrieval.

\subsubsection{Ingredient embedding}
Regarding the ingredient embedding process, we implement LSTM and transformer \cite{vaswani2017attention} respectively as the encoder to produce ingredient features $\mathrm{F}_{ing}$. We first follow \cite{salvador2017learning,wang2019learning} to use the ingredient-level word2vec representations $y \in \{y_1, \dots, y_n\}$ for ingredient token representations. For example, \emph{ground ginger} is regarded as a single word vector, instead of two separate word vectors of \emph{ground} and \emph{ginger}. Then the processed word2vec representations $y$ are fed into the ingredient encoder, where we experiment with both the bidirectional LSTM and the transformer. Specifically, we follow \cite{vaswani2017attention,wang2021cross} and implement the self-attention mechanism on the LSTM to boost the performance for the fair comparison with the transformer. The transformer is constructed with 4 layers. We use the final state output of the ingredient encoder as the ingredient features $\mathrm{F}_{ing}$.

\subsubsection{Cooking instruction embedding}
To obtain the cooking instruction features $\mathrm{F}_{ins}$, we also follow previous practice \cite{salvador2017learning,wang2019learning} to extract the sentence features for fair comparisons. We first obtain the fixed-length representation $r \in \{r_1, \dots, r_n\}$ for each cooking instruction sentence with the skip-thoughts \cite{kiros2015skip} technique. Then we feed $r$ into the instruction encoder to get the sequence embeddings $\mathrm{F}_{ins}$ for cooking instructions. Here we also experiment with both the LSTM and the transformer as the instruction encoder, where the LSTM is enhanced with the self-attention mechanism \cite{vaswani2017attention} and the transformer has 4 layers. 

\subsubsection{Tree structure embedding}
We further newly introduce the sentence-level tree structure representations to improve the cooking instruction features. To this end, we produce the tree structures $T$ from the given cooking instruction sentences, which are generated by the recipe2tree module introduced in Section \ref{sec:r2t}. $T$ is converted into the adjacency matrix $A$, such that we can use GAT to emb sentence-level trees $T$ and obtain structure representations $\mathrm{F}_{tree}$ for cooking instructions. It is notable that here tree representation $\mathrm{F}_{tree}$ construction method is different from that in Section \ref{sec:tree2re}, where the cooking instruction sentence representations are not available during generation phase, in the retrieval setting we can use sentence embeddings as node features. Technically, we denote the cooking instruction sentence embeddings from the skip-thoughts \cite{kiros2015skip} as $r \in \{r_1, \dots, r_n\}$. The child node representations are set as $r$, and the parent node representation are set as the mean of its child node representations. Hence the node representations with the sentence embeddings can be denoted as $f_{node}^{sen}$. Moreover, since the learned tree structures have the hierarchical property, we incorporate additional embeddings $f_{node}^{depth}$ on the node depth, such that the learned tree representations include both the node relationships and the node hierarchy. The node input node features $f^{node}$ are constructed by the concatenation of $f^{node}_{sen}$ and $f^{node}_{depth}$.

Therefore, we can compute the attention coefficients as below:
\begin{equation}
    e_{ij}^{node} = f^{node}_i {f^{node}_j}^T,
\end{equation}
where we use the matrix multiplication to measure the relationships between the node features $(f^{node}_i, f^{node}_j)$. Eq. (\ref{eq: alpha}) is further adopted to give the attentional scores $\alpha_{ij}$. With $\alpha_{ij}$, we can formulate $\mathrm{F}_{tree}$ as
\begin{equation}
    \mathrm{F}_{tree} = \sigma(\sum_{j \in \mathrm{N}_i} \alpha_{ij}f^{node}_j),
\end{equation}
where $\mathrm{N}_i$ denotes the neighborhood of node $i$, and $\sigma$ is the nonlinear activation used in GAT.

\subsubsection{Retrieval training}
The recipe representations $\mathrm{F}_{rec}$ is obtained from the concatenation of the ingredient features $\mathrm{F}_{ing}$, instruction features $\mathrm{F}_{ins}$ and recipe tree structure representation $\mathrm{F}_{tree}$.
We utilize the triplet loss to train image-to-recipe retrieval model, the objective function is:
\begin{equation}
\begin{aligned}
L_{tri} =  & \sum\left[d(\mathrm{F}_{img}^{a}, \mathrm{F}_{rec}^{p})-d(\mathrm{F}_{img}^{a}, \mathrm{F}_{rec}^{n})+m\right] \; \\
& + \sum\left[d(\mathrm{F}_{rec}^{a}, \mathrm{F}_{img}^{p})-d(\mathrm{F}_{rec}^{a}, \mathrm{F}_{img}^{n})+m\right],
\end{aligned}
\end{equation}
where $d(\bullet)$ denotes the Euclidean distance, superscripts $a,p$ and $n$ refer to anchor, positive and negative samples respectively and $m$ is the margin. We follow the practice of previous works \cite{wang2019learning,wang2021cross} to use the \emph{BatchHard} idea proposed in \cite{hermans2017defense}, to improve the training effectiveness. Specifically, we dynamically construct the triplets during training phase. In a mini-batch, we select the most distant positive instance and the closest negative instance, given an anchor sample.

\section{Experiments} \label{sec:exp}

\subsection{Dataset and evaluation metrics}
We evaluate our proposed method on Recipe1M dataset \cite{salvador2017learning}, which is one of the largest collection of cooking recipe data with food images. Recipe1M has rich food related information, including the food images, ingredients and cooking instructions. In Recipe1M, there are $252,547$, $54,255$ and $54,506$ food data samples for training, validation and test respectively. These recipe data is collected from some public websites, which are uploaded by users.

For the recipe generation task, we evaluate the model using the same metrics as prior works \cite{salvador2019inverse,wang2020structure}: perplexity, BLEU \cite{papineni2002bleu} and ROUGE \cite{lin2004rouge}. Perplexity is used in \cite{salvador2019inverse}, it measures how well the learned word probability distribution matches the target recipes. BLEU is computed based on the average of unigram, bigram, trigram and 4-gram precision. We use ROUGE-L to test the longest common subsequence. ROUGE-L is a modification of BLEU, where ROUGE-L score metric is measuring recall instead of precision. Therefore, we can use ROUGE-L to measure the fluency of generated recipes.

Regrading the image-to-recipe retrieval task, we evaluate our proposed framework as the common practice used in prior works \cite{salvador2017learning,chen2018deep,carvalho2018cross,zhu2019r2gan,wang2019learning}. To be specific, median retrieval rank (MedR) and recall at top K (R@K) are used. MedR measures the median rank position among where true positives are returned. Therefore, higher performance comes with a lower MedR score. Given a food image, R@K calculates the fraction of times that the correct recipe is found within the top-K retrieved candidates, and vice versa. Different from MedR, the performance is directly proportional to the score of R@K. In the test phase, we first sample 10 different subsets of 1,000 pairs (1k setup), and 10 different subsets of 10,000 (10k setup) pairs. It is the same setting as in \cite{salvador2017learning}. We then consider each item from food image modality in subset as a query, and rank samples from recipe modality according to L2 distance between the embedding of image and that of recipe, which is served as image-to-recipe retrieval, and vice versa for recipe-to-image retrieval.

\begin{table*}
  \centering
    \caption{Food cross-modal retrieval evaluation. We show results of baseline models and our proposed method. The models are evaluated on the basis of MedR (lower is better), and R@K (higher is better).}
  \begin{tabular}{cl|cccccccc}
  \toprule
   \textbf{Size of Test Set} && \multicolumn{4}{c}{\textbf{Image-to-Recipe Retrieval}} & \multicolumn{4}{c}{\textbf{Recipe-to-Image Retrieval} }\\
    \midrule
    \multicolumn{1}{c}{} & \textbf{Methods} & \textbf{medR $\downarrow$} & \textbf{R@1 $\uparrow$} &\textbf{R@5 $\uparrow$}& \textbf{R@10 $\uparrow$} &  \textbf{medR $\downarrow$} & \textbf{R@1 $\uparrow$} & \textbf{R@5 $\uparrow$}& \textbf{R@10 $\uparrow$} \\
    \midrule
	\multirow{9}{*}{{1k}} 
        &CCA ~\cite{hotelling1936relations}& 15.7 & 14.0 & 32.0 & 43.0 & 24.8 & 9.0 & 24.0 & 35.0 \\
        &SAN ~\cite{chen2017cross}& 16.1 & 12.5 & 31.1 & 42.3 & - & - & - & -\\
        &JE ~\cite{salvador2017learning}& 5.2 & 24.0 & 51.0 & 65.0 & 5.1 & 25.0 & 52.0 & 65.0  \\
        &AM ~\cite{chen2018deep}& 4.6 & 25.6 & 53.7 & 66.9 & 4.6 & 25.7 & 53.9 & 67.1  \\
        &AdaMine  ~\cite{carvalho2018cross}& \textbf{1.0} & 39.8 & 69.0 & 77.4 & \textbf{1.0} & 40.2 & 68.1 & 78.7 \\
        &$\rm {R^2 GAN}$ ~\cite{zhu2019r2gan} & \textbf{1.0} & 39.1 & 71.0 & 81.7 & \textbf{1.0} & 40.6 & 72.6 & 83.3 \\
        &ACME ~\cite{wang2019learning} & \textbf{1.0} & 51.8 & 80.2 & 87.5 & \textbf{1.0} & 52.8 & 80.2 & 87.6 \\
        &Ours & \textbf{1.0} & \textbf{53.5}  & \textbf{81.5}  & \textbf{88.8}  & \textbf{1.0} & \textbf{55.0}  & \textbf{82.0}  & \textbf{88.8} \\
    \midrule
	\multirow{6}{*}{{10k}} 
	& JE ~\cite{salvador2017learning} & 41.9 & - & - & - & 39.2 & - & - & - \\
	&AM ~\cite{chen2018deep}& 39.8 & 7.2 & 19.2 & 27.6 & 38.1 & 7.0 & 19.4 & 27.8  \\ 
    &AdaMine  ~\cite{carvalho2018cross}& 13.2 & 14.9 & 35.3 & 45.2 & 12.2 & 14.8 & 34.6 & 46.1\\
    &$\rm {R^2 GAN}$ ~\cite{zhu2019r2gan} & 13.9 & 13.5 & 33.5 & 44.9 & 11.6 & 14.2 & 35.0 & 46.8 \\
    &ACME ~\cite{wang2019learning} & 6.7 & 22.9 & 46.8 & 57.9 & 6.0 & 24.4 & 47.9 & 59.0 \\
    &Ours & \textbf{6.0} & \textbf{23.4} & \textbf{48.8}  & \textbf{60.1}  & \textbf{5.6} & \textbf{24.6}  & \textbf{50.0}  & \textbf{61.0} \\
    \bottomrule
  \end{tabular}
  \label{tab:retri}
\end{table*}

\subsection{Implementation details}
We adopt a $3$-layer ON-LSTM \cite{shen2018ordered} to output the sentence-level tree structure, taking about $50$ epoch training to get converged. We set the learning rate as $1$, batch size as $60$, and the input embedding size is $400$, which is the same as original work \cite{shen2018ordered}. We select recipes containing over $4$ sentences in Recipe1M dataset for training. And we randomly select several consecutive sentences as the context and the following one as the correct one. We set $K$ as $3$. We show some of the predicted sentence-level tree structures for recipes in Figure \ref{fig:parsing_vis}. 

We use two different ingredient encoders in the experiments, i.e. the non-pretrained and pretrained language model. Using non-pretrained model is to compare with the prior work \cite{salvador2019inverse}, where they use a word embedding layer to give the ingredient embeddings. We use BERT \cite{devlin2018bert} as the pretrained language model, giving $512$-dimensional features. The image encoder is used with a ResNet-50 \cite{he2016deep} pretrained on ImageNet \cite{deng2009imagenet}. And we map image output features to the dimension of $512$, to align with the ingredient features. We adopt a RNN for tree adjacency sequence generation, where the RNN initial hidden state is initialized as the previous image features. The RNN layer is set as $2$ and the hidden state size is $512$. The tree embedding model is graph attention network (GAT), its attention head number is set as $6$. The output tree feature dimension is set the same as that of image features. We use the same settings in language decoder as prior work \cite{salvador2019inverse}, a $16$-layer transformer \cite{vaswani2017attention}. The number of attention heads in the decoder is set as $8$. We use greedy search during text generation, and the maximum generated instruction length is $150$. We set $\lambda_1$ and $\lambda_2$ in Eq. (\ref{eq:obj}) as $1$ and $0.5$ respectively. The model is trained using Adam \cite{kingma2014adam} optimizer with the batch size of $16$. Initial learning rate is set as $0.001$, which decays $0.99$ each epoch. The BERT model finetune learning rate is $0.0004$.

In the retrieval model training, we use the pretrained ResNet-50 model to give image features. We then adopt an one-layer bi-directional LSTM with the self-attention mechanism \cite{vaswani2017attention} and a 4-layer transformer respectively to encode the recipes, to show the difference between using LSTM and transformer in the food retrieval task. An $8$-head GAT is used to encode the sentence-level tree structures to give the tree features, which are concatenated with the recipe features. We map the image and recipe features to a common space to do the retrieval training, with the feature size of $1024$. We set the batch size and learning rate as $64$ and $0.0001$ respectively. We decrease the learning rate $0.1$ in the 30th epoch.  

In terms of the computation cost, we observe our introduced components do not add too much burden to the original model. We use a single V100 for all experiments. The recipe2tree module is shared by the recipe generation and cross-modal retrieval models, where its training phase costs about 4 hours and inference time is about 0.01 second for each sample. For the recipe generation model, the baseline model \cite{devlin2018bert} takes around 1.5 hours per epoch for training and 0.04 second per sample for inference; the baseline model \cite{devlin2018bert} with structural representations takes around 1.7 hours per epoch for training and 0.05 second per sample for inference.
For the food cross-modal retrieval model, the baseline model (LSTM) takes around 0.5 hour per epoch for training and 0.02 second per sample for inference; the baseline model (LSTM) using structural representations takes around 0.6 hour per epoch for training and 0.02 second per sample for inference.

\subsection{Baselines}

\subsubsection{Recipe generation}
Since Recipe1M has different data components from standard MS-COCO dataset \cite{chen2015microsoft}, it is hard to implement some prior image captioning models in Recipe1M. To the best of our knowledge, \cite{salvador2019inverse} is the only recipe generation work on Recipe1M dataset, where they use the Encoder-Decoder architecture. Based on the ingredient and image features, they generate the recipes with transformer \cite{vaswani2017attention}. 

The SGN model we proposed is an extension of the baseline model, which learns the sentence-level tree structure of target recipes by an unsupervised approach. We infer the tree structures of recipes before language generation, adding an additional module on the baseline model. It means that our proposed SGN can be applied to many other deep model architectures and vision-language datasets. We test the performance of SGN with two ingredient encoders, 1) non-pretrained word embedding model and 2) pretrained BERT model. Word embedding model is used in \cite{salvador2019inverse}, trained from scratch. BERT model \cite{devlin2018bert} is served as another baseline, to test if SGN can improve language generation performance further under a powerful encoder. We use ResNet-50 in both two baseline models. The recipe generation results are shown in Table \ref{tab:main}.

\subsubsection{Food cross-modal retrieval}
Canonical Correlation Analysis (CCA) \cite{hotelling1936relations} is one of the most widely-used classic models for learning a common embedding from different feature spaces, which learns linear projections for images and text to maximize their feature correlation. Salvador et al. \cite{salvador2017learning} aim to learn joint embeddings (JE) for images and recipes, where they adopt cosine loss to align image-recipe pairs and classification loss to regularize the learning. In SAN \cite{chen2017cross} and AM \cite{chen2018deep}, they introduce attention mechanism to different levels of recipes including food titles, ingredients and cooking instructions. AdaMine \cite{carvalho2018cross} is an adaptive learning schema in the training phase, helping the model perform an adaptive mining for significant triplets. Later, adversarial methods \cite{zhu2019r2gan,wang2019learning} are proposed for retrieval alignment. Specifically, Zhu et al. \cite{zhu2019r2gan} use two-level ranking loss at embedding and image spaces in $\mathbf{R^2 GAN}$. ACME \cite{wang2019learning} introduce the translation consistency component to allow feature distributions from different modalities to be similar. The evaluation results are shown in Table \ref{tab:retri}.

\begin{table}
% % \vspace{-0.1in}
  \centering
  \caption{Recipe generation evaluation. We give results of baseline models without and with the proposed SGN for comparison. We also present ablative results of the pretrained model performance without the image and ingredient (ingr) features respectively. The model is evaluated with perplexity, BLEU and ROUGE-L.}
  {
%   % \vspace{-0.1in}
\resizebox{\columnwidth}{!}{
    \begin{tabular}{l|ccc}
    \toprule
    % \cline{1-4}
     \textbf{Methods}  & \textbf{Perplexity $\downarrow$} & \textbf{BLEU $\uparrow$} & \textbf{ROUGE-L $\uparrow$}\\
    \midrule
    \textbf{Non-pretrained Model \cite{salvador2019inverse}} & 8.06  & 7.23  & 31.8  \\
    \textbf{\quad\quad\quad\quad\quad\quad\quad\quad\quad\quad + SGN} &   7.46  & 9.09  & 33.4 \\
    % \cline{1-4}
    % \cline{1-4}
    \midrule
    \textbf{TIRG \cite{vo2019composing}} & 7.83   &  7.95 & 32.4 \\
    \textbf{\quad\quad\quad\quad\quad\quad\quad\quad\quad\quad + SGN} &  7.56   &  9.24 & 34.5 \\
    \textbf{VAL \cite{chen2020image}} & 7.61   &  8.83 & 34.2 \\
    \textbf{\quad\quad\quad\quad\quad\quad\quad\quad\quad\quad + SGN} &  7.05   &  10.43 & 35.6 \\
    \midrule
    \textbf{Pretrained Model \cite{devlin2018bert}} & 7.52 & 9.29  & 34.8 \\
    \textbf{\quad\quad\quad\quad\quad\quad\quad\quad\quad\quad - ingr} &  8.16   &  3.72 & 31.0 \\
    \textbf{\quad\quad\quad\quad\quad\quad\quad\quad\quad\quad - image} &  7.62   &  5.74 & 32.1 \\
    \textbf{\quad\quad\quad\quad\quad\quad\quad\quad\quad\quad + SGN} &  \textbf{6.67} & \textbf{12.75}  & \textbf{36.9} \\
    % \cline{1-4}
    \bottomrule
    \end{tabular}}%
  }
  \label{tab:main}%
% % \vspace{-0.2in}
\end{table}%

\subsection{Evaluation results}

\subsubsection{Recipe generation}
\noindent \textbf{Language generation performance.} We show the performance of SGN for recipe generation against the baselines in Table \ref{tab:main}. In both baseline settings, our proposed method SGN outperforms the baselines across all metrics. In the method of non-pretrained model, SGN achieves a BLEU score more than $9.00$, which is about $25\%$ higher than the current state-of-the-art method. Here we directly concatenate the image and text features. To compare the impact of different image-text features fusion methods, we also give results of TIRG \cite{vo2019composing} and VAL \cite{chen2020image}. Specifically, TIRG \cite{vo2019composing} adopts an LSTM and a ResNet CNN to encode the text and the images respectively, then with the gating and residual connections the image-text fused features can be obtained. 
In VAL \cite{chen2020image}, Chen et al. also use an LSTM and a ResNet-50 to get the text and image features respectively. They feed the concatenation of the image and text features into the transformer, where the concatenated features are further processed with the attention mechanism to produce the fused features. We observe there is a margin between the performance of TIRG and VAL, since VAL use the composite transformers that are plugged into different convolution layers to compose the vision and language contents, obtaining more fine-grained features than TIRG. 

When we shift to the pretrained model method \cite{devlin2018bert}, we can see that the pretrained language model gets comparable results as ``TIRG + SGN" model.
We also show the ablative results of models trained without ingredient (ingr) and image features respectively, where we observe the ingredient features help more on the generation results. When incrementally adding SGN to the pretrained model, the performance of SGN is significantly superior to all the baselines by a substantial margin. Although we only use the concatenation method to fuse the image and text features, we utilize the pretrained BERT model to extract the text features, which gives better results than ``VAL + SGN". This may indicate the significance of the pretrained language model.
On the whole, the efficacy of SGN is shown to be very promising, outperforming the state-of-the-art method across different metrics consistently.

\begin{table}
  \centering
  \caption{Generated recipe average length for recipe generation. We compare the average length between recipes from different sources.}
  {
    \begin{tabular}{l|cc}
    \toprule
     \textbf{Methods}  & \textbf{Recipe Average Length}\\
    \midrule
    \textbf{Pretrained Model \cite{devlin2018bert}} & 66.9  \\
    \textbf{\quad\quad\quad\quad\quad\quad\quad\quad\quad + SGN} &   112.5 \\

    \midrule
    \textbf{Ground Truth (Human)} & 116.5 \\

    \bottomrule
    \end{tabular}%
  }
  \label{tab:length}%
\end{table}%

\begin{table}\centering
\caption{Ablation studies on the composition of tree features for cross-modal retrieval. We adopt different tree node embeddings and report the results on rankings of size $1k$, with the basis of R@K (higher is better). Here we use the LSTM to encode the recipes.}\label{tab: ab_fea}
\begin{tabular}{l|ccc}\toprule
\textbf{Node Features} &\textbf{R@1 $\uparrow$} & \textbf{R@5 $\uparrow$} & \textbf{R@10 $\uparrow$} \\\midrule
\textbf{Adjacency matrix projection} & 52.7 & 81.3 & 88.4 \\
\textbf{$f_{node}^{sen}$} & 53.3 & 81.5 & 88.5 \\
\textbf{$f_{node}^{sen}$ + $f_{node}^{depth}$} & \textbf{53.5}  & \textbf{81.5}  & \textbf{88.8} \\
\bottomrule
\end{tabular}
\end{table}

\noindent \textbf{Impact of structure awareness.} To explicitly suggest the impact of tree structures on the final recipe generation, we compute the average length for the generated recipes, as shown in Table \ref{tab:length}. Average length can reflect the text structure on node numbers. It is observed that SGN generates recipes with the most similar length as the ground truth, indicating the help of the tree structure awareness.

\begin{table}\centering
\caption{Ablation studies of different recipe encoders for cross-modal retrieval, where the LSTM and transformer are used respectively as the backbone to encode the cooking recipes. We experiment with various methods to fuse the tree and recipe features, where \textbf{cat} represents the concatenation method, TIRG \cite{vo2019composing} and VAL \cite{chen2020image} are the existing cross-modal feature learning methods. The results are reported on rankings of size $1k$, with the basis of R@K (higher is better).
}\label{tab: retri_ab}
\begin{tabular}{l|ccc}\toprule
\textbf{Method} & \textbf{R@1 $\uparrow$} & \textbf{R@5 $\uparrow$} & \textbf{R@10 $\uparrow$} \\\midrule
\textbf{baseline (LSTM)}  &52.5 &81.1 &88.4 \\
\quad\quad\quad\quad\quad\quad\quad\quad \textbf{+ tree (cat)} &53.5 & 81.5 & 88.8\\
\quad\quad\quad\quad\quad\quad\quad\quad \textbf{+ tree (TIRG \cite{vo2019composing})} &54.2 & 81.9 & 88.6 \\
\quad\quad\quad\quad\quad\quad\quad\quad \textbf{+ tree (VAL \cite{chen2020image})} & 54.0 & \textbf{82.0} & 89.0 \\\midrule
\textbf{baseline (transformer)} & 53.4 & 81.5 & 87.8 \\
\quad\quad\quad\quad\quad\quad\quad\quad \textbf{+ tree (cat)} & 54.3 & 81.6 & 88.4 \\
\quad\quad\quad\quad\quad\quad\quad\quad \textbf{+ tree (TIRG \cite{vo2019composing})} & \textbf{54.7} & 81.9 & \textbf{89.1} \\
\quad\quad\quad\quad\quad\quad\quad\quad \textbf{+ tree (VAL \cite{chen2020image})} & 54.4 & 81.8 & 88.7 \\
\bottomrule
\end{tabular}
\end{table}

\subsubsection{Food cross-modal retrieval}
\noindent \textbf{Cross-modal retrieval performance.} In Table \ref{tab:retri}, we compare the results of our proposed method with various state-of-the-art methods against different metrics. Specifically, ACME \cite{wang2019learning} takes triplet loss and adversarial training to learn image-recipe alignment, which gives superior results over other state-of-the-art models. ACME mainly focuses on improving cross-modal representation consistency at the common space with the cross-modal translation. Here we do not use the adversarial training of ACME and implement the self-attention mechanism on the LSTM to train the retrieval model. We add the tree structure representations on the baseline, it can be observed that we further boost the performance across all the metrics. It suggests that our unsupervisedly-learned tree structures also can be applied on the retrieval task and have positive value to the whole model.  

\noindent \textbf{Ablation study of different compositions of tree features.} In Table \ref{tab: ab_fea}, we adopt the LSTM as the recipe encoder. We first directly use the projection of the adjacency matrix as the node features, and further construct the tree embeddings with GAT. We then adopt the sentence features $f_{node}^{sen}$ and the concatenation of $f_{node}^{sen}$ and $f_{node}^{depth}$ respectively to be the node features. We can see that the including more information into the tree features helps on improving the food cross-modal retrieval performance.

\noindent \textbf{Ablation study of different recipe encoders.} We show the results of using LSTM and transformer respectively as the recipe embedding backbone in Table \ref{tab: retri_ab}. We adopt the self-attention based LSTM \cite{vaswani2017attention} and transformer to encode the recipes. It can be seen that since the attention mechanism is implemented on the LSTM, the LSTM model achieves similar performance to the transformer model. And the learned tree features can boost the retrieval performance in both settings.

\begin{table}\centering
\caption{Ablation studies of incorporating the img2tree module into the cross-modal retrieval model. \emph{recipe2tree} indicates using the tree structure for recipe features only. \emph{img2tree} represents we also learn to generate the tree structures from the images during the retrieval training phase. The \emph{coefficient $\lambda$} indicates trade-off parameter on the tree generation loss $L_{tree}$. 
The results are reported on rankings of size $1k$, with the basis of R@K (higher is better).}\label{tab: retri_img2tree}
\begin{tabular}{l|c|ccc}\toprule
\textbf{Method} & \textbf{coefficient $\lambda$} & \textbf{R@1 $\uparrow$} & \textbf{R@5 $\uparrow$} & \textbf{R@10 $\uparrow$} \\\midrule
\textbf{recipe2tree only} &  & \textbf{53.5} & \textbf{81.5} & \textbf{88.8} \\\midrule
\textbf{\quad\quad + img2tree} & 1 & 52.5 & 81.1 & 88.4 \\
\textbf{\quad\quad + img2tree} & 0.1 & 53.4 & 81.4 & 88.5 \\
\textbf{\quad\quad + img2tree} & 0.01 & 52.9 & 81.3 & 88.4 \\
\bottomrule
\end{tabular}
\end{table}

\noindent \textbf{Ablation study of different feature fusion methods.} Since we use the concatenation of the ingredient features $\mathrm{F}_{ing}$, cooking instruction features $\mathrm{F}_{ins}$ and recipe tree structure features $\mathrm{F}_{tree}$ as the recipe representations, here we also adopt TIRG \cite{vo2019composing} and VAL \cite{chen2020image} respectively to replace the concatenation method and produce the fused features of $\mathrm{F}_{ing}$, $\mathrm{F}_{ins}$ and $\mathrm{F}_{tree}$. The experiment results are shown in Table \ref{tab: retri_ab}. Technically, TIRG \cite{vo2019composing} utilized the gating and residual connections to obtain the cross-modal fused features. VAL \cite{chen2020image} applied spatial and channel attention transform on the image-text features.
Note that here we do not fuse the image and text features together, but the tree and text features instead. That means we cannot apply the spatial self-attention transformation on the tree-text features as the original implementation in VAL \cite{chen2020image}. Therefore, we only conduct the channel attention of VAL \cite{chen2020image}. The results show TIRG \cite{vo2019composing} gives more improvements over the baseline than VAL \cite{chen2020image}. We also observe that adding tree features or using better feature fusion methods mainly brings improvements on the R@1, which means better top-ranking results can be obtained.

\noindent \textbf{Ablation study of using the img2tree module.} In Table \ref{tab: retri_img2tree}, we aim to evaluate the performance of incorporating the img2tree module into the retrieval framework. Specifically, we learn to generate the trees from food images, which is supervised by the pseudo ground truth from the recipe2tree module. train the model with loss of $L_{tri} + \lambda L_{tree}$, where $\lambda$ is the trade-off parameter. The inferred tree structures are also encoded with the GAT and concatenated with the food image features, which are used as the query to retrieve the corresponding recipes. We conduct experiments with various $\lambda$. However, we observe there is no performance improvement of adding the img2tree module on the image feature learning, which may indicate the differences between the recipe generation and food cross-modal retrieval tasks. Technically, in the retrieval setting, we attempt to do precise matching between the image and recipe features, where the image features contain the generated tree representations and recipe features contain the pseudo ground truth tree representations. If the predicted tree structures from the images are not accurate enough, i.e. fail to produce the same node numbers and the same node relationships as the pseudo ground truth, then the tree embeddings will also be different, which decreases the retrieval performance. While in the recipe generation task, the extra guidance of plausible tree structures can help the model capture the preliminary cooking procedure, thus giving better performance than the model without inferred structures. Nevertheless, better results can also be expected if the prediction performance of img2tree module can be improved further. We leave it for the future work.

\begin{figure}
\begin{center}
\includegraphics[width=0.45\textwidth]{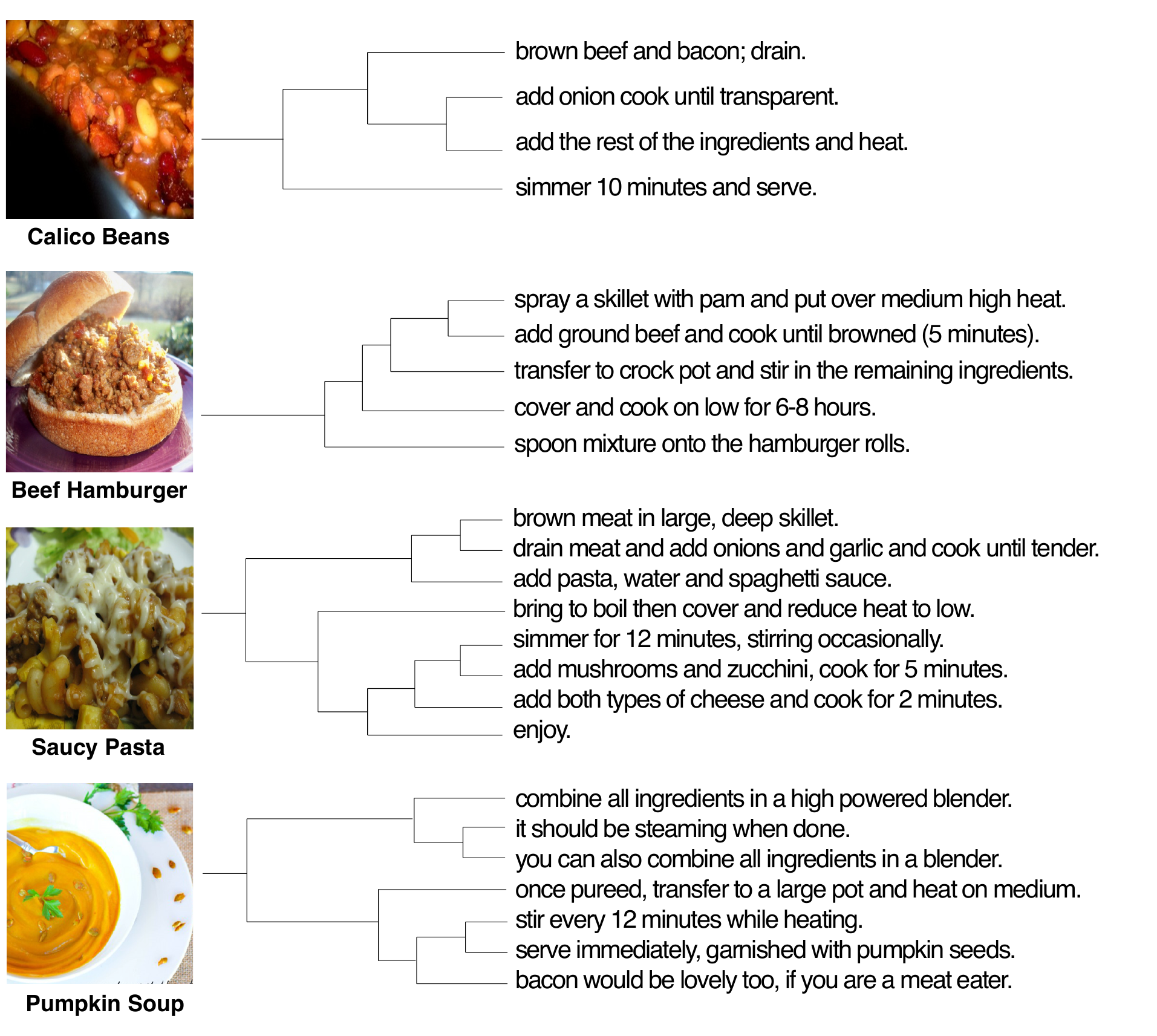}
\end{center}
% \vspace{-0.2in}
   \caption{The visualization of predicted sentence-level trees for recipes. The latent tree structure is obtained from unsupervised learning. The results indicate that we can get reasonable parsing tree structures with varying recipe length.}
\label{fig:parsing_vis}
% \vspace{-0.2in}
\end{figure}

\begin{figure*}
\begin{center}
\includegraphics[width=0.8\textwidth]{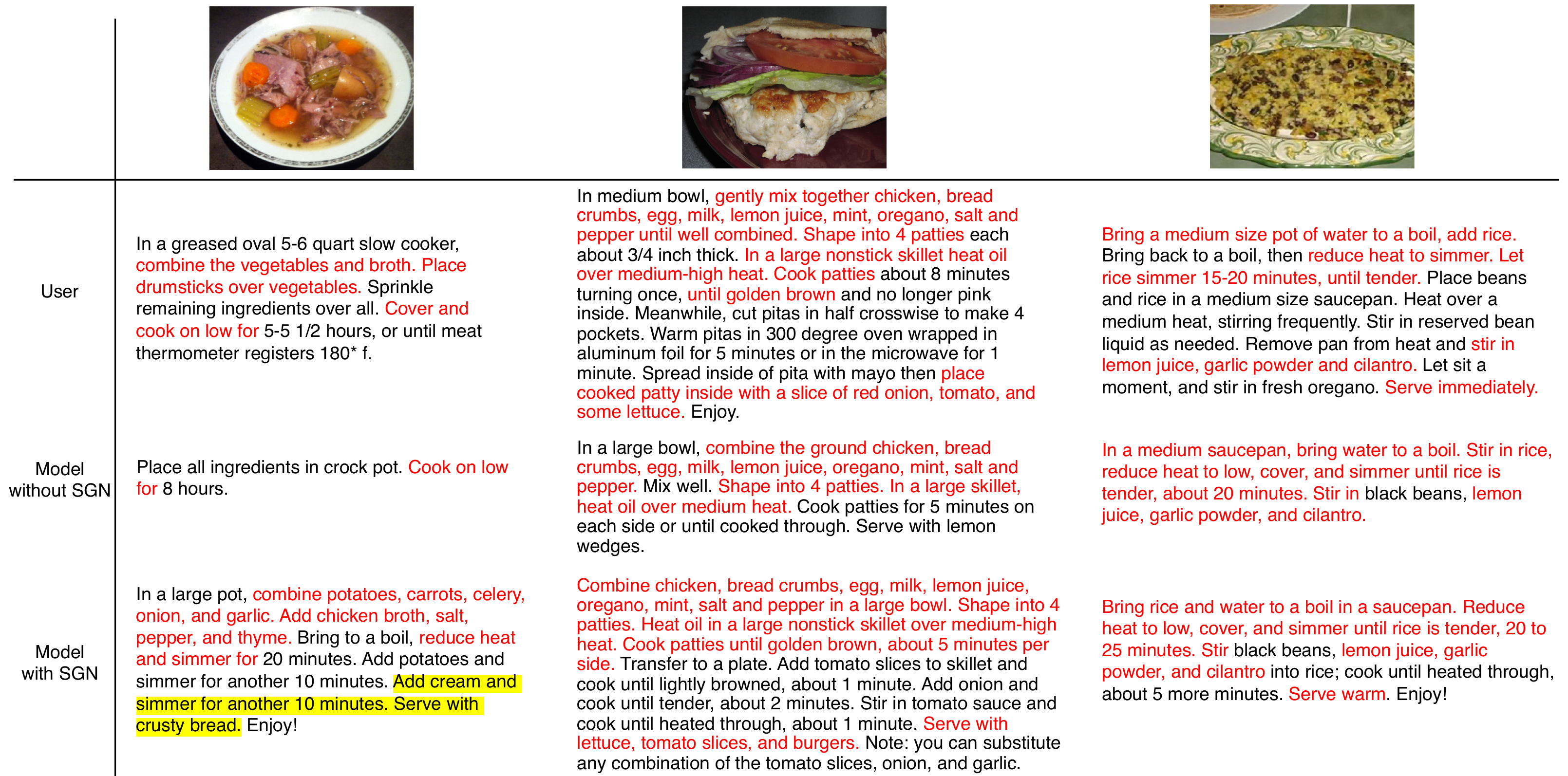}
\end{center}
% \vspace{-0.2in}
   \caption{Visualization of recipes from different sources. We show the food images and the corresponding recipes, obtained from users and different types of models. Words in red indicate the matching parts between recipes uploaded by users and that generated by models. Words in yellow background show the redundant generated sentences.}
\label{fig:gen_vis}
% \vspace{-0.1in}
\end{figure*}

\begin{figure*}
\begin{center}
\includegraphics[width=0.7\textwidth]{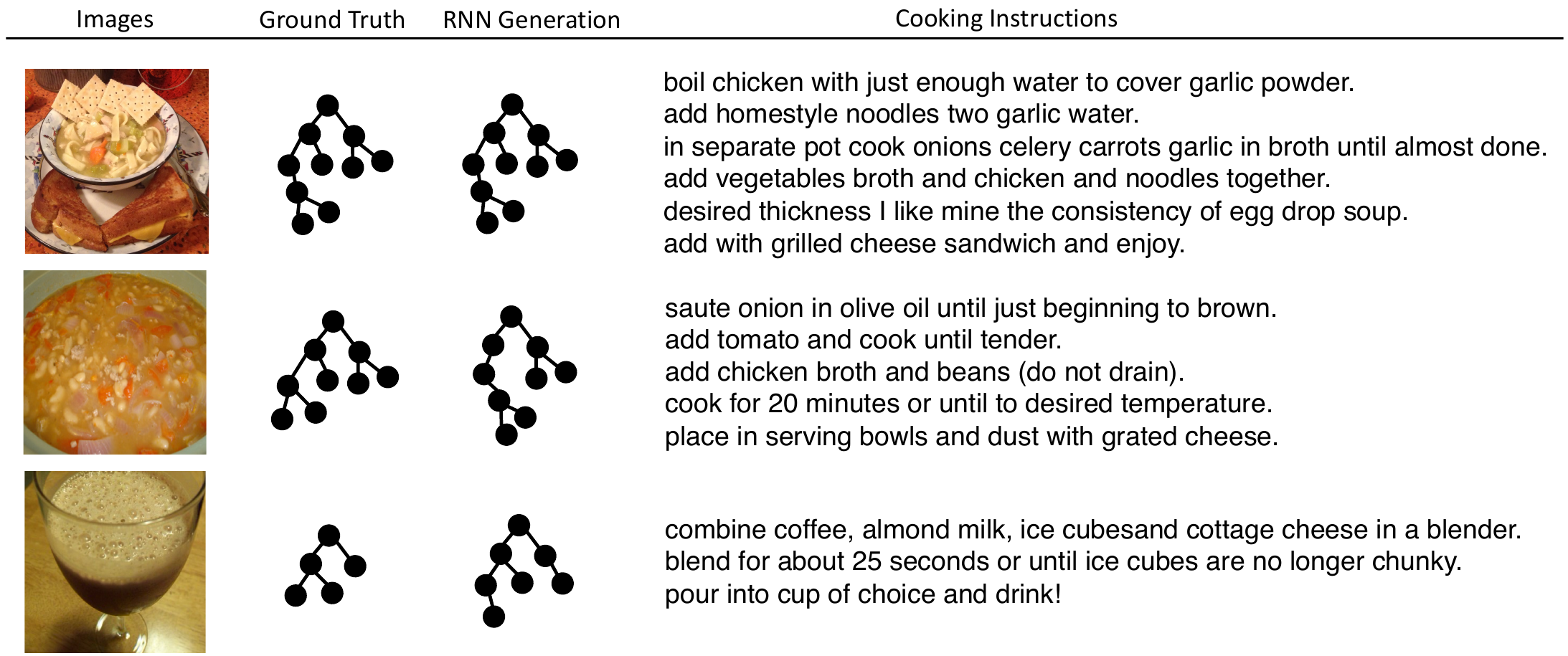}
\end{center}
% \vspace{-0.2in}
   \caption{The comparison between the pseudo ground truth trees (produced by recipe2tree module) and img2tree generated tree structures.}
\label{fig:graph_vis}
% \vspace{-0.2in}
\end{figure*}

\subsection{Qualitative results}

\subsubsection{Sentence-level tree parsing results}
In Figure \ref{fig:parsing_vis}, we visualize some parsing tree results of our proposed recipe2tree module. Due to there is no human labelling on the recipe tree structures, we can hardly provide a quantitative analysis on the parsing trees. 

We show some examples with varying paragraph length in Figure \ref{fig:parsing_vis}. The first two rows show the tree structures of relatively short recipes. Take the first row (\emph{calico beans}) as example, the generated tree set the food pre-processing part (step 1) as a separate leaf node, and two main cooking steps (step 2\&3) are set as deeper level nodes. The last \emph{simmer} step is conditioned on previous three steps, which is put in another different tree level. We can see that the parsing tree results correspond with common sense and human experience. 

In the last two rows of Figure \ref{fig:parsing_vis}, we show the parsing results of recipes having more than $5$ sentences. The tree of \emph{pumpkin soup} indicates clearly two main cooking phases, i.e. before and after ingredient pureeing. Generally, the proposed recipe2tree generated sentence-level parsing trees look plausible, helping on the inference for recipe generation.

\begin{figure*}
\begin{center}
\includegraphics[width=0.7\textwidth]{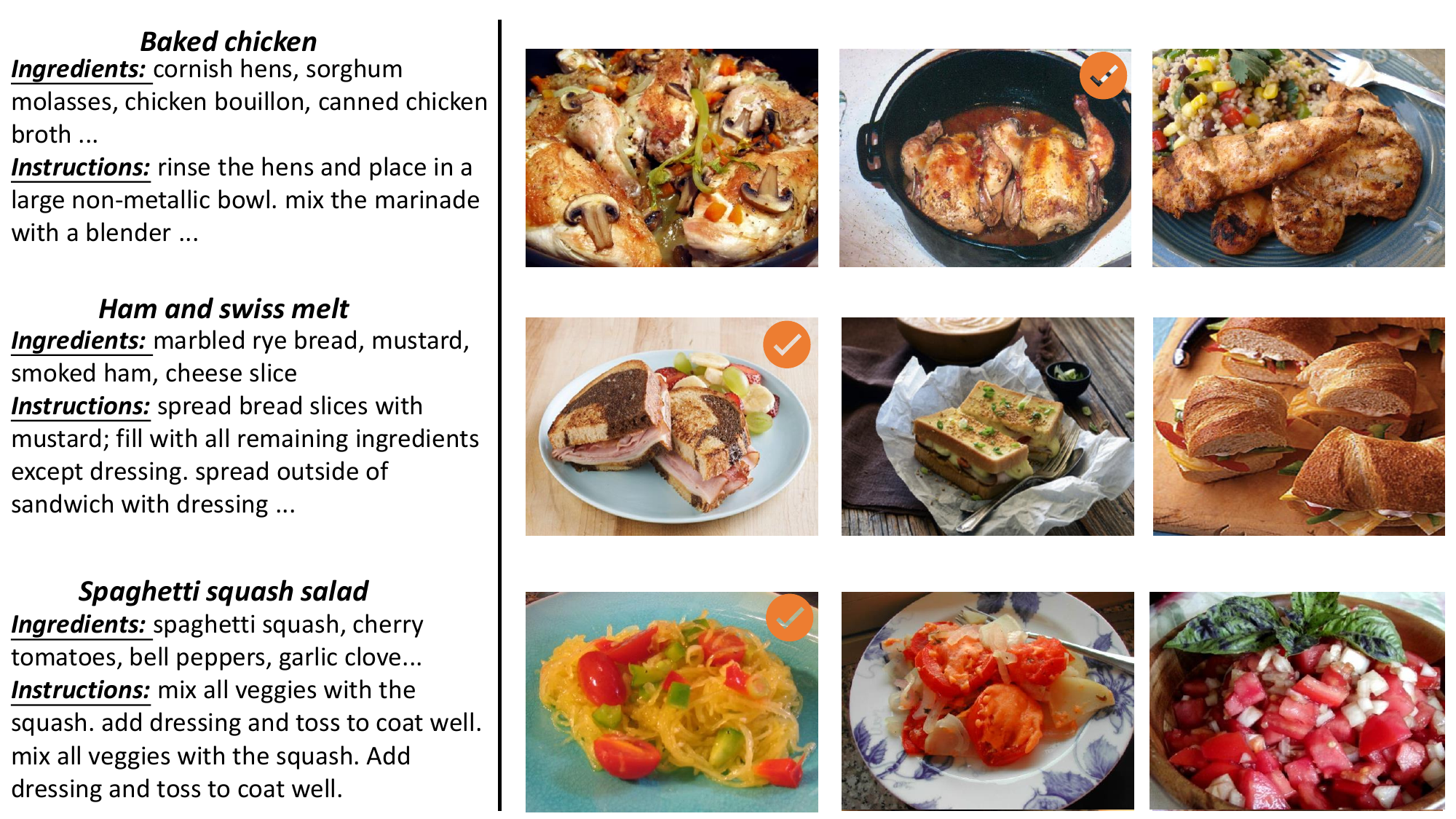}
\end{center}
% \vspace{-0.2in}
   \caption{The visualization of recipe-to-image retrieval results. For each recipe query, we show the top-3 retrieved food images, where the yellow ticks suggest the correctly retrieved samples.}
\label{fig:re2im}
% \vspace{-0.2in}
\end{figure*}

\subsubsection{Recipe generation results}
We present some recipe generation results in Figure \ref{fig:gen_vis}. We consider three types of recipe sources, the human, models trained without and with SGN. Each recipe accompanies with a food image. We can observe that recipes generated by model with SGN have similar length with that written by users. It may indicate that, instead of generating language directly from the image features, allowing the deep model to be aware of the structure first brings benefits for the following recipe generation task. 

We indicate the matching parts between recipes provided by users and that generated by models, in red words. It is observed that SGN model can produce more coherent and detailed recipes than non-SGN model. For example, in the middle column of Figure \ref{fig:gen_vis}, SGN generated recipes include some ingredients that do not exist in the non-SGN generation, but are contained in users' recipes, such as \emph{onion}, \emph{lettuce} and \emph{tomato}.

However, although SGN can generate longer recipes than non-SGN model, it may produce some redundant sentences. These useless sentences are marked with yellow background, as shown in the first column of Figure \ref{fig:gen_vis}. Since \emph{Cream} is not supposed to be used in the chicken soup, in the future work, we may need to use the input ingredient information better to guide the recipe generation.

\subsubsection{Tree generation results}
There are some graph evaluation metrics proposed in \cite{you2018graphrnn}, however, these metrics are used for unconditional graph evaluation. How to evaluate the graph similarities for conditional generation remains an open problem. Here we show some examples of generated recipe tree structures in Figure \ref{fig:graph_vis} for qualitative analysis. Tree generation results from image features are by-product of our proposed SGN framework. They are used to improve the final recipe generation performance.  

It is notable that only the leaf nodes in the tree represent the sentences of recipe. We can observe that the overall img2tree generated structures look similar with the ground truth trees, which are produced by recipe2tree module. And the generated trees have some diversity. However, it is hard to align the number of generated nodes with the ground truth. For example, in the last row of Figure \ref{fig:graph_vis}, the generated tree has one more node than the ground truth. 
Nevertheless, the generated trees along with the tree embedding network can improve the final recipe generation performance based on the quantitative results. 

\begin{figure}
\begin{center}
\includegraphics[width=0.45\textwidth]{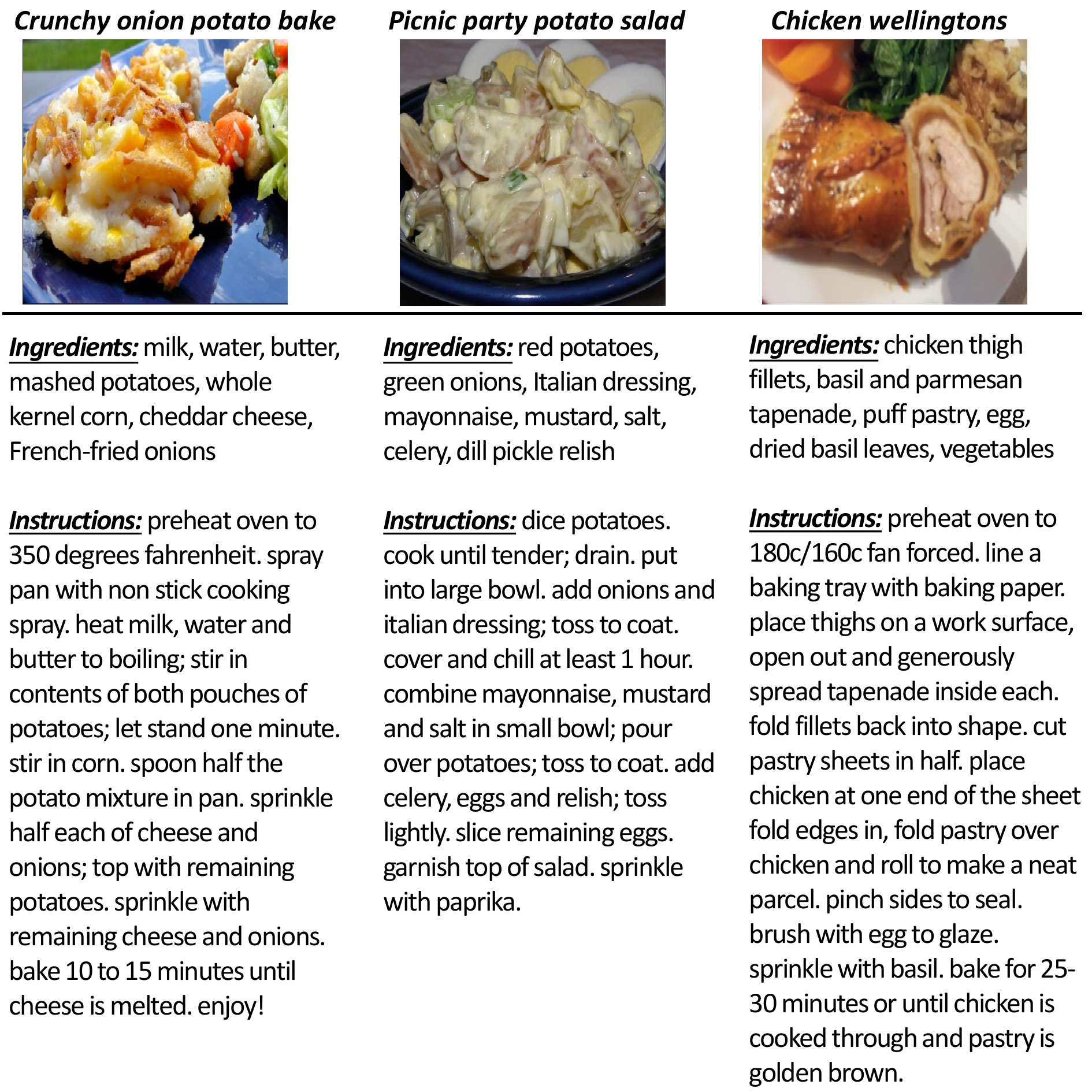}
\end{center}
% \vspace{-0.2in}
   \caption{The visualization of image-to-recipe retrieval results. The top row indicates the image query, the bottom row shows the corresponding retrieved recipes, which are correctly matched with the ground truth.}
\label{fig:im2re}
% \vspace{-0.2in}
\end{figure}

\subsubsection{Food cross-modal retrieval results}
We show the visualizations of recipe-to-image and image-to-recipe retrieval in Figure \ref{fig:re2im} and \ref{fig:im2re} respectively, where we apply 1k setup. In Figure \ref{fig:re2im}, we conduct the recipe-to-image retrieval. Specifically, we also select three different recipes and take them as queries, the right column lists the top-3 retrieved results. The results demonstrate that our model not only has good performance on cross-modal retrieval, and also helps to discover similar samples from the same modality. For example, in the top row all of our top-3 ranked food images contain \emph{chicken}, which share semantic similarity. In Figure \ref{fig:im2re}, we randomly select images from three different food classes, i.e. \emph{potato cake}, \emph{salad} and \emph{chicken wellingtons}. Based on the image queries, we can see that our model retrieves the cooking recipes across various food classes correctly. This also indicates some potential applications, for example, the model can automatically produce the corresponding cooking recipes given users' uploaded food images.

\section{Conclusion}
In this paper, we have proposed a novel unsupervised learning approach to generate the sentence-level tree structures for the cooking recipes, which can benefit the recipe generation and food cross-modal retrieval tasks. To be specific, 
we propose effective ways to address some challenging problems, including unsupervisedly extracting the paragraph structures, generating tree structures from images and using the tree structures for recipe generation and food cross-modal retrieval models. Technically, we extend ON-LSTM to label recipe tree structures using an unsupervised manner. In the recipe generation model, we propose to use RNN to generate the tree structures from food images, and adopt the inferred trees to enhance the generation. In the cross-modal retrieval model, we integrate the tree structural information into the cooking recipe representations.
We demonstrate our learned tree structures improve both of the recipe generation and cross-modal model performance. We have conducted quantitative and qualitative analysis for both tasks, the results show that models with our proposed tree structures outperform various baseline models.

However, since the food image semantic structures can hardly be obtained, we attempt to learn the structural information for text only in our proposed framework, while we fail to build the inner semantic relationships for visual data. In the future work, we can try to construct the structural representations for the visual data. We may extend our current work to cooking video datasets, where the relationships between video frames and the correspondence between recipes and frames can be built.

\ifCLASSOPTIONcompsoc
  \section*{Acknowledgments}
\else
  \section*{Acknowledgment}
\fi
This research is supported, in part, by the National Research Foundation (NRF), Singapore under its AI Singapore Programme (AISG Award No: AISG-GC-2019-003) and under its NRF Investigatorship Programme (NRFI Award No. NRF-NRFI05-2019-0002). Any opinions, findings and conclusions or recommendations expressed in this material are those of the authors and do not reflect the views of National Research Foundation, Singapore. This research is supported, in part, by the Singapore Ministry of Health under its National Innovation Challenge on Active and Confident Ageing (NIC Project No. MOH/NIC/HAIG03/2017).
This research is also supported by the National Research Foundation, Singapore under its AI Singapore Programme (AISG Award No: AISG-RP-2018-003), and the MOE AcRF Tier-1 research grant: RG95/20.

\ifCLASSOPTIONcaptionsoff
  \newpage
\fi

% \newpage



\bibliographystyle{IEEEtran}
\bibliography{egbib}

% \newpage


\begin{IEEEbiography}[{\includegraphics[width=1in,height=1.25in,clip,keepaspectratio]{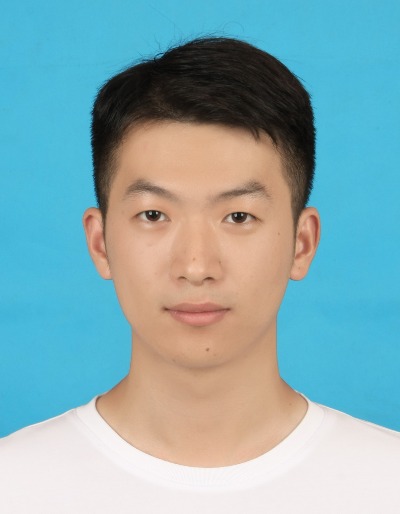}}]{Hao Wang}
is a PhD candidate with the School of Computer Science and Engineering, Nanyang Technological University, Singapore. His research interests include cross-modal generation and computer vision.
\end{IEEEbiography}
% \begin{IEEEbiographynophoto}{Hao Wang}
% is a PhD candidate with the School of Computer Science and Engineering, Nanyang Technological University, Singapore. His research interests include multi-modal analysis and computer vision.

%\newpage

\begin{IEEEbiography}[{\includegraphics[width=1in,height=1.25in,clip,keepaspectratio]{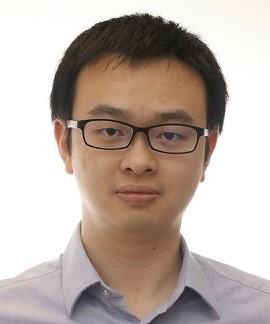}}]{Guosheng Lin} 
is currently an Assistant Professor with the School of Computer Science and Engineering, Nanyang Technological University, Singapore. His research interests include computer vision and machine learning.
\end{IEEEbiography}
% \begin{IEEEbiographynophoto}{Guosheng Lin} 
% is currently an Assistant Professor with the School of Computer Science and Engineering, Nanyang Technological University, Singapore. His research interests include computer vision and machine learning.

\begin{IEEEbiography}[{\includegraphics[width=1in,height=1.25in,clip,keepaspectratio]{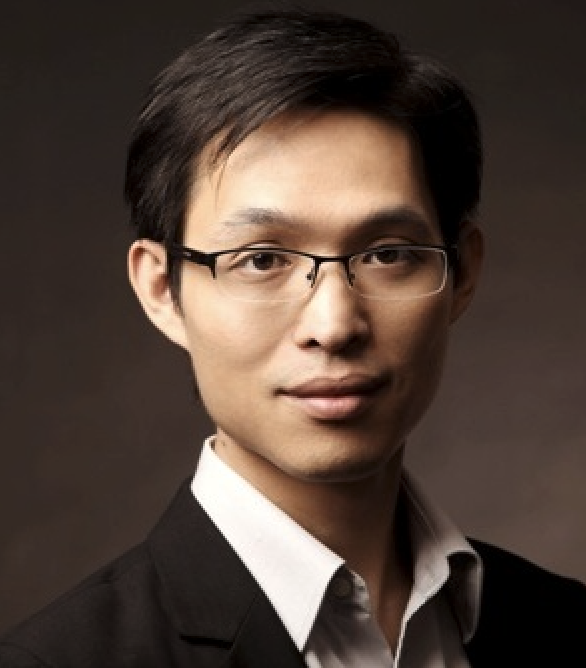}}]{Steven C. H. Hoi}
is currently the Managing Director of Salesforce Research Asia, and a Professor of Information Systems at Singapore Management University, Singapore. He received his Bachelor degree from Tsinghua University, P.R. China, in 2002, and his Ph.D degree in computer science and engineering from The Chinese University of Hong Kong, in 2006. He has served as the Editor-in-Chief for Neurocomputing Journal, general co-chair for ACM SIGMM Workshops on Social Media, program co-chair for the fourth Asian Conference on Machine Learning, book editor for “Social Media Modeling and Computing”, guest editor for ACM Transactions on Intelligent Systems and Technology. He is an IEEE Fellow and ACM Distinguished Member.
\end{IEEEbiography}

\begin{IEEEbiography}[{\includegraphics[width=1in,height=1.25in,clip,keepaspectratio]{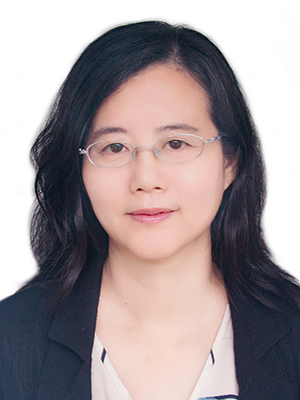}}]{Chunyan Miao} 
is the chair of School of Computer Science and Engineering in Nanyang Technological
University (NTU), Singapore. Dr. Miao is a Full Professor
in the Division of Information Systems and Director of
the Joint NTU-UBC Research Centre of Excellence in
Active Living for the Elderly (LILY), School of Computer
Engineering, Nanyang Technological University (NTU),
Singapore. She received her PhD degree from NTU and
was a Postdoctoral Fellow/Instructor in the School of
Computing, Simon Fraser University (SFU), Canada. She
visited Harvard and MIT, USA, as a Tan Chin Tuan Fellow, collaborating on a large NSF funded research program in social networks and virtual worlds. She has been an Adjunct Associate
Professor/Associate Professor/Founding Faculty member with the Center for Digital
Media which is jointly managed by The University of British Columbia (UBC) and
SFU. She is the Editor-in-Chief of the International Journal of Information Technology published by the Singapore Computer
Society.
\end{IEEEbiography}


%\vfill


\end{document}